\documentclass[10pt,twocolumn,letterpaper]{article}

\usepackage{wacv}
\usepackage{times}
\usepackage{epsfig}
\usepackage{graphicx}
\usepackage{amsmath}
\usepackage{amssymb}
\usepackage{bbding}
\usepackage{balance}
\wacvfinalcopy

\ifwacvfinal
\usepackage[breaklinks=true,bookmarks=false]{hyperref}
\else
\usepackage[pagebackref=true,breaklinks=true,colorlinks,bookmarks=false]{hyperref}
\fi

\begin{document}

\title{The MECCANO Dataset: Understanding Human-Object Interactions from \\
Egocentric Videos in an Industrial-like Domain}

\author{Francesco Ragusa\\
IPLAB, University of Catania\\
XGD-XENIA s.r.l., Acicastello, Catania, Italy\\
{\tt\small francesco.ragusa@unict.it}\\
\\

Salvatore Livatino\\
University of Hertfordshire\\
{\tt\small s.livatino@herts.ac.uk}

\and
Antonino Furnari\\
IPLAB, University of Catania\\
{\tt\small furnari@dmi.unict.it}
\\
\\
\\
Giovanni Maria Farinella\\
IPLAB, University of Catania\\
{\tt\small gfarinella@dmi.unict.it}

}

\maketitle

\begin{abstract}
Wearable cameras allow to collect images and videos of humans interacting with the world. 
While human-object interactions have been thoroughly investigated in third person vision, the problem has been understudied in egocentric settings and in industrial scenarios. To fill this gap, we introduce MECCANO, the first dataset of egocentric videos to study human-object interactions in industrial-like settings. 
MECCANO has been acquired by $20$ participants who were asked to build a motorbike model, for which they had to interact with tiny objects and tools. 
The dataset has been explicitly labeled for the task of recognizing human-object interactions from an egocentric perspective. Specifically, each interaction has been labeled both temporally (with action segments) and spatially (with active object bounding boxes).
With the proposed dataset, we investigate four different tasks including 1) action recognition, 2) active object detection, 3) active object recognition and 4) egocentric human-object interaction detection, which is a revisited version of the standard human-object interaction detection task.
Baseline results show that the MECCANO dataset is a challenging benchmark to study egocentric human-object interactions in industrial-like scenarios. 
We publicy release the dataset at \url{https://iplab.dmi.unict.it/MECCANO}.
\end{abstract}

\section{Introduction}

\begin{figure}[t]
\centering
\includegraphics[width=\columnwidth]{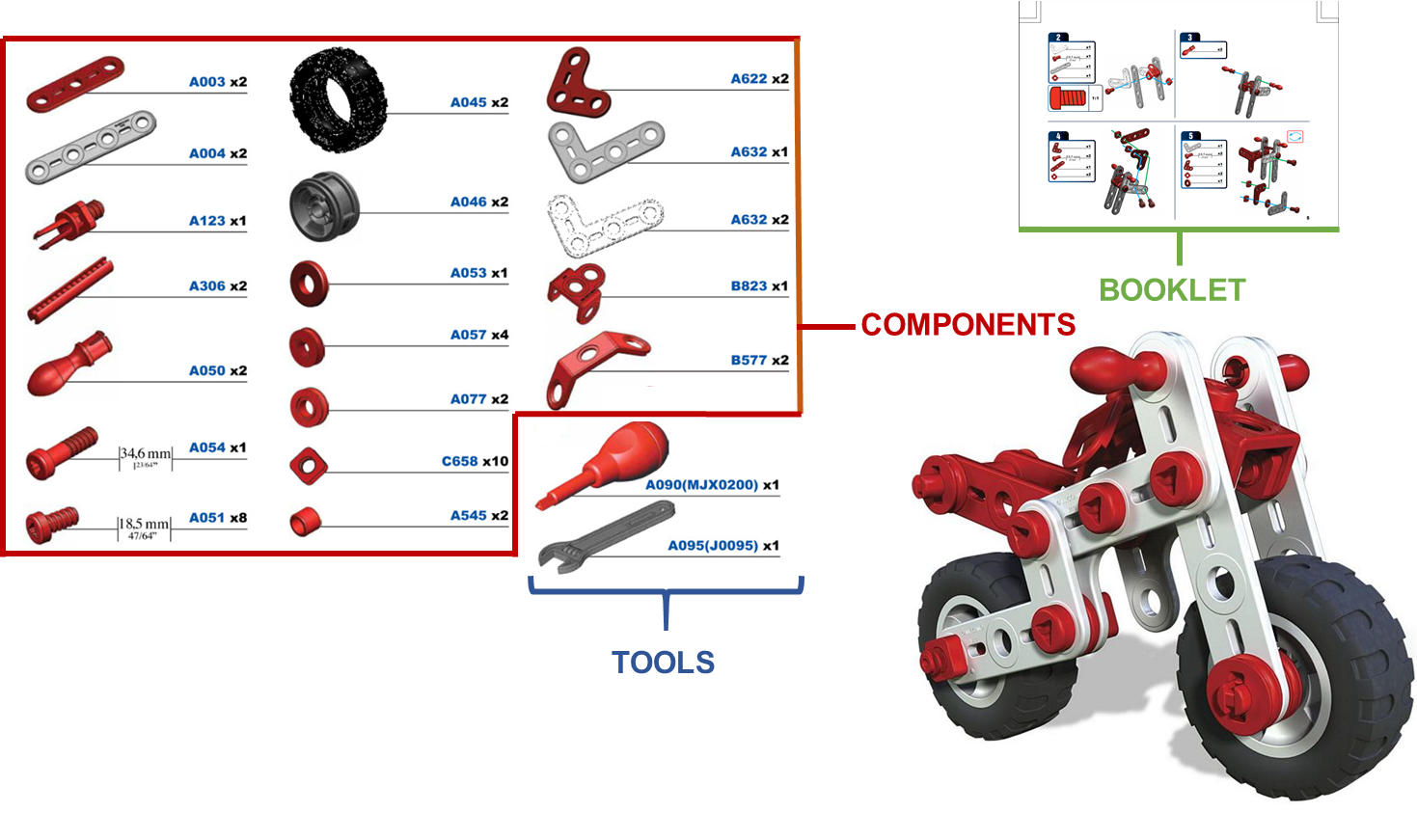}
\caption{Toy model built by subjects interacting with 2 tools, 49 components and the instructions booklet. Better seen on screen.}
\label{fig:toy_model}
\end{figure}

Being able to analyze human behavior from egocentric observations has many potential applications related to the recent development of wearable devices \cite{Epson_moverio, Holo2, Vuzix} which range from improving the personal safety of workers in a factory~\cite{DeepVisionShield_Colombo19} to providing assistance to the visitors of a museum~\cite{vedi2019, RagusaPRL, cucchiara2014visions}.
In particular, with the rapid growth of interest in wearable devices in industrial scenarios, recognizing human-object interactions can be useful to prevent safety hazards, implement energy saving policies and issue notifications about actions that may be missed in a production pipeline \cite{miss_actions_shapiro}. 

In recent years, progress has been made in many research areas related to human behavior  understanding, such as action recognition \cite{feichtenhofer2018slowfast, TwoStream_convolutional_action_Zisserman_14, Two-Stream_Zisserman, Zhou2018TemporalRR, kazakos2019TBN, TSM_2019}, object detection \cite{girshick2014rich, girshick2015fast, ren2015faster, yolov3} and human-object interaction detection \cite{Gkioxari2018DetectingAR, Gupta2015VisualSR, Hands_in_contact_Shan20, Nagarajan2020EGOTOPOEA}. These advances have been possible thanks to the availability of large-scale datasets \cite{Imagenet, lin2014COCO, Damen2018EPICKITCHENS, Li2018_EGTEA-GAZE+, Gupta2015VisualSR, HICO_Chao} which have been curated and often associated with dedicated challenges. In the egocentric vision domain, in particular, previous investigations have considered the contexts of kitchens \cite{Damen2018EPICKITCHENS, Li2018_EGTEA-GAZE+, Torre2009CMU-MMAC}, as well as daily living activity at home and in offices \cite{Ramanan_12_ADL, thu-read_17, You-Do_Damen_14, Ortis2017OrganizingEV}. While these contexts provide interesting test-beds to study user behavior in general, egocentric human-object interactions have not been previously studied in industrial environments such as factories, building sites, mechanical workshops, etc. This is mainly due to the fact that data acquisition in industrial domains is difficult because of privacy issues and the need to protect industrial secrets. \\

In this paper, we present MECCANO, the first dataset of egocentric videos to study human-object interactions in industrial-like settings. To overcome the limitations related to data collection in industry, we resort to an industrial-like scenario in which subjects are asked to build a toy model of a motorbike using different components and tools (see Figure~\ref{fig:toy_model}). Similarly to an industrial scenario, the subjects interact with tools such as a screwdriver and a wrench, as well as with tiny objects such as screws and bolts while executing a task involving sequential actions (e.g., take wrench, tighten bolt, put wrench). Despite the fact that this scenario is a simplification of what can be found in real industrial settings, it is still fairly complex, as our experiments show.

\begin{figure}[t]
\centering
\includegraphics[width=0.48\textwidth]{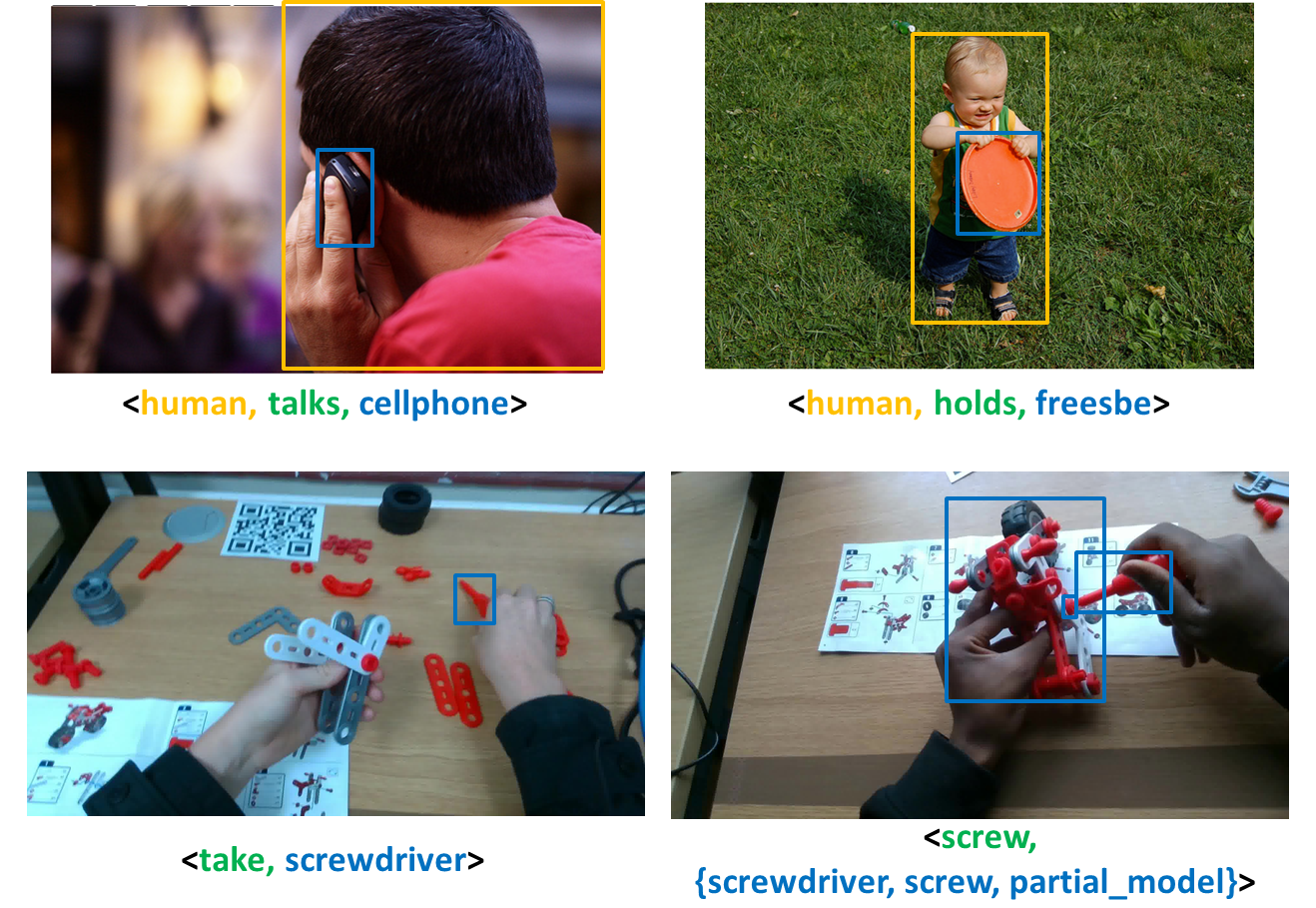}
\caption{Examples of Human-Object Interactions in third person vision (first row) and first person vision (second row)\protect\footnotemark.}
\label{fig:concept}
\end{figure}

MECCANO was collected by 20 different participants in two countries (Italy and United Kingdom). We densely annotated the acquired videos with temporal labels to indicate the start and end times of each human-object interaction, and with bounding boxes around the active objects involved in the interactions. We hope that the proposed dataset will encourage research in this challenging domain. The dataset is publicly released at the following link: \url{https://iplab.dmi.unict.it/MECCANO}.

We show that the proposed dataset can be used to study four fundamental tasks related to the understanding of human-object interactions: 1)~Action Recognition, 2)~Active Object Detection, 3)~Active Object Recognition and 4)~Egocentric Human-Object Interaction Detection. While past works have already investigated the tasks of action recognition \cite{Damen2018EPICKITCHENS, Li2018_EGTEA-GAZE+, Ma_2016_CVPR, Sudhakaran_2019_CVPR}, active object detection \cite{Ramanan_12_ADL}, and active object recognition~\cite{Damen2018EPICKITCHENS} in the context of egocentric vision, Human-Object Interaction (HOI) detection has been generally studied in the context of third person vision \cite{Gupta2015VisualSR, Gkioxari2018DetectingAR, Qi2018LearningHI, Chao2018LearningTD, RPN_Zhou, PPDM_liao2019, Wang_InteractionPoints_2020_CVPR}. Since we believe that modelling actions both semantically and spatially is fundamental for egocentric vision applications, we instantiate the Human-Object Interaction detection task in the context of the proposed dataset.

HOI detection consists in detecting the occurrence of human-object interactions, localizing both the humans taking part in the action and the interacted objects. HOI detection also aims to understand the relationships between humans and objects, which is usually described with a verb. 
Possible examples of HOIs are  \textit{``talk on the cell phone''} or \textit{``hold a fresbee''}. HOI detection models mostly consider one single object involved in the interaction \cite{Gupta2015VisualSR, HOI_Gupta_09, Gkioxari2018DetectingAR, HOI_Fei_Fei,Chao2018LearningTD}. Hence, an interaction is generally defined as a triplet in the form \textit{$<$human, verb, object$>$}, where the human is the subject of the action specified by a verb and an object. Sample images related to human-object interactions in a third-person scenario are shown in Figure~\ref{fig:concept}-top. 
We define Egocentric Human-Object Interaction (EHOI) detection as the task of producing $<$verb, objects$>$ pairs describing the interaction observed from the egocentric point of view. Note that in EHOI, the human interacting with the objects is always the camera wearer, while one or more objects can be involved simultaneously in the interaction. The goal of EHOI detection is to infer the verb and noun classes, and to localize each active object involved in the interaction. Figure~\ref{fig:concept}-bottom reports some examples of Egocentric Human-Object Interactions. 

We perform experiments with baseline approaches to tackle the four considered tasks. Results suggest that the proposed dataset is a challenging benchmark for understanding egocentric human-object interactions in industrial-like settings.
In sum, the contributions of this work are as follows: 1) we present MECCANO, a new challenging egocentric dataset to study human-object interactions in an industrial-like domain; 2) we instantiate the HOI definition in the context of egocentric vision (EHOI); 3) we show that the current state-of-the-art approaches achieve limited performance, which suggests that the proposed dataset is an interesting benchmark for studying egocentric human-object interactions in industrial-like domains.

\begin{table*}[]
\resizebox{\textwidth}{!}{%
\setlength\tabcolsep{2pt}
\begin{tabular}{lccclcccccccc}
\multicolumn{1}{c}{\textbf{Dataset}} & \textbf{Settings} & \textbf{EGO?} & \textbf{Video?} & \multicolumn{1}{c}{\textbf{Tasks}} & \textbf{Year} & \textbf{Frames} & \textbf{Sequences} & \textbf{AVG. video duration} & \textbf{Action classes} & \textbf{Object classes} & \textbf{Object BBs} & \textbf{Participants} \\ \hline
MECCANO & Industrial-like & \checkmark & \checkmark & EHOI, AR, AOD, AOR & 2020 & 299,376 & 20 & 20.79 min & 61 & 20 & 64,349 & 20 \\ \hline
EPIC-KITCHENS \cite{Damen2018EPICKITCHENS} & Kitchens & \checkmark & \checkmark & AR, AOR & 2018 & 11,5M & 432 & 7.64 min & 125 & 352 & 454,255 & 32 \\
EGTEA Gaze+ \cite{Li2018_EGTEA-GAZE+} & Kitchens & \checkmark & \checkmark & AR & 2018 & 2,4M & 86 & 0.05 min & 106 & 0 & 0 & 32 \\
THU-READ \cite{thu-read_17} & Daily activties & \checkmark & \checkmark & AR & 2017 & 343,626 & 1920 & 7.44 min & 40 & 0 & 0 & 8 \\
ADL \cite{Ramanan_12_ADL} & Daily activities & \checkmark & \checkmark & AR, AOR & 2012 & 1,0M & 20 & 30.0 min & 32 & 42 & 137,780 & 20 \\
CMU \cite{Torre2009CMU-MMAC} & Kitchens & \checkmark & \checkmark & AR & 2009 & 200,000 & 16 & 15.0 min & 31 & 0 & 0 & 16 \\ \hline
Something-Something \cite{Something_Something_Goyal} & General & X & \checkmark & AR, HOI & 2017 & 5,2 M & 108,499 & 0.07 min & 174 & N/A & 318,572 & N/A \\
Kinetics \cite{Kinetics_Carreira2019ASN} & General & X & \checkmark & AR & 2017 & N/A & 455,000 & 0.17 min & 700 & 0 & 0 & N/A \\
ActivityNet \cite{caba2015activitynet} & Daily activites & X & \checkmark & AR & 2015 & 91,6 M & 19,994 & 2.55 min & 200 & N/A & N/A & N/A \\ \hline
HOI-A \cite{PPDM_liao2019} & General & X & X & HOI, AOR & 2020 & 38,668 & N/A & N/A & 10 & 11 & 60,438 & N/A \\
HICO-DET \cite{HICO_Chao} & General & X & X & HOI, AOR & 2018 & 47,776 & N/A & N/A & 117 & 80 & 256,672 & N/A \\
V-COCO \cite{Gupta2015VisualSR} & General & X & X & HOI, OD & 2015 & 10,346 & N/A & N/A & 26 & 80 & N/A & N/A \\ \hline
\end{tabular}%
}
\caption{Comparative overview of relevant datasets. HOI: HOI Detection. EHOI: EHOI Detection. AR: Action Recognition. AOD: Active Object Detection. AOR: Active Object Recognition. OD: Object Detection.}
\label{tab:datasets}
\end{table*}

\footnotetext{Images in the first row were taken from the COCO dataset \cite{lin2014COCO} while those in the second row belong to the MECCANO dataset.}

\section{Related Work}
\label{sec:related_work}
\textbf{Datasets for Human Behavior Understanding \hspace{1mm}} Previous works have proposed datasets to tackle the task of Human-Object Interaction (HOI) detection.
Among the most notable datasets, we can mention V-COCO~\cite{Gupta2015VisualSR}, which adds $26$ verb labels to the $80$ objects of COCO~\cite{lin2014COCO}, HICO-Det \cite{HICO_Chao}, labeled with $117$ verbs and $80$ objects, HOI-A~\cite{PPDM_liao2019}, which focuses on $10$ verbs and $11$ objects indicating actions dangerous while driving. Other works have proposed datasets for action recognition from video. Among these, ActivityNet \cite{caba2015activitynet} is a large-scale dataset composed of videos depicting $203$ activities that are relevant to how humans spend their time in their daily lives, Kinetics~\cite{Kinetics_2017, Kinetics_Carreira2019ASN} is a dataset containing $700$ human action classes, Something-Something~\cite{Something_Something_Goyal} includes low-level concepts to represent simple everyday aspects of the world. 
Previous works also proposed datasets of egocentric videos. Among these datasets, EPIC-Kitchens~\cite{Damen2018EPICKITCHENS, Damen2020RESCALING, Damen2020Collection} focuses on unscripted activities in kitchens, EGTEA Gaze+~\cite{Li2018_EGTEA-GAZE+} contains videos paired with gaze information collected from participants cooking different recipes in a kitchen, CMU~\cite{Torre2009CMU-MMAC} is a multi-modal dataset of egocentric videos including RGB, audio and motion capture information, ADL~\cite{Ramanan_12_ADL} contains egocentric videos of subjects performing daily activities, THU-READ~\cite{thu-read_17} contains RGB-D videos of subjects performing daily-life actions in different scenarios.
Table~\ref{tab:datasets} compares the aforementioned datasets with respect to the proposed dataset. MECCANO is the first dataset of egocentric videos collected in an industrial-like domain and annotated to perform EHOI Detection. 
It is worth noting that previous egocentric datasets have considered scenarios related to kitchens, offices, and daily-life activities and that they have generally tackled the action recognition task rather than EHOI detection.

\begin{figure*}[t]
\centering
\includegraphics[width=0.9\textwidth]{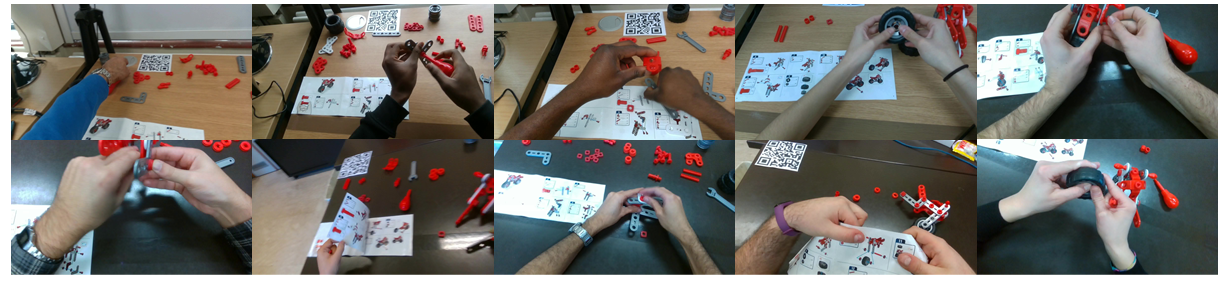}
\caption{Examples of data acquired by the 20 different participants in two countries (Italy, United Kingdom).}
\label{fig:dataset}
\end{figure*}

\textbf{Action Recognition \hspace{1mm}} Action recognition from video has been thoroughly investigated especially in the third person vision domain. Classic works~\cite{Learning_actions_movies_Laptev, Human_detection_Flow_Schmid_06} relied on motion-based features such as optical flow and space-time features.
Early deep learning works fused processing of RGB and optical flow features with two-stream networks~\cite{TwoStream_convolutional_action_Zisserman_14, Two-Stream_Zisserman, temporal_segNet}, 3D ConvNets are commonly used to encode both spatial and temporal dimensions~\cite{ Conv_spatio-temporal_Taylor, Learning_spatio-temporal_Paluri, Carreira2017QuoVA}, long-term filtering and pooling has focused on representing actions considering their full temporal extent \cite{Long-term_action_Schmid, Two-Stream_Zisserman, temporal_segNet, Zhou2018TemporalRR,Zhou2018TemporalRR,TSM_2019}. 
Other works separately factor convolutions into separate 2D spatial and 1D temporal filters~\cite{Spatiotemporal_residual_action, closer_spatiotemp_action, rethinking_spatiotemporal, Learning_spatiotemporal_pseudo}.
Among recent works, Slow-Fast networks~\cite{feichtenhofer2018slowfast} avoid using pre-computed optical flow and encodes motion into a ``fast'' pathway (which operates at a high frame rate) and simultaneously a ``slow'' pathway which captures semantics (operating at a low frame rate). 
We asses the performance of state-of-the-art action recognition methods on the proposed dataset considering SlowFast networks~\cite{feichtenhofer2018slowfast}, I3D~\cite{Carreira2017QuoVA} and 2D CNNs as baselines.

\textbf{HOI Detection \hspace{1mm}} Previous works have investigated HOI detection mainly from a third person vision point of view. 

The authors of~\cite{Gupta2015VisualSR} proposed a method to detect people performing actions able to localize the objects involved in the interactions on still images.
The authors of~\cite{Gkioxari2018DetectingAR} proposed a human-centric approach based on a three-branch architecture (InteractNet) instantiated according to the definition of HOI in terms of a $<$human, verb, object$>$ triplet. 
Some works~\cite{Qi2018LearningHI, Chao2018LearningTD, RPN_Zhou} explored HOI detection using graph convolutional neural networks after detecting humans and objects in the scene.
Recent works~\cite{PPDM_liao2019, Wang_InteractionPoints_2020_CVPR} represented the relationship between both humans and objects as the intermediate point which connects the center of the human and object bounding boxes.
The aforementioned works addressed the problem of HOI detection in the third person vision domain. In this work, we look at the task of HOI detection from an egocentric perspective considering the proposed MECCANO dataset.

\textbf{EHOI Detection \hspace{1mm}}
EHOI detection is understudied due to the limited availability of egocentric datasets labelled for this task. 
While some previous datasets such as EPIC-KITCHENS~\cite{Damen2018EPICKITCHENS,Damen2020Collection} and ADL~\cite{Ramanan_12_ADL} have been labeled with bounding box annotations, these datasets have not been explicitly labeled for the EHOI detection task indicating relationships between labeled objects and actions, hence preventing the development of EHOI detection approaches.
Some related studies have modeled the relations between entities for interaction recognition as object affordances~\cite{Hotspots_Grauman19, Nagarajan2020EGOTOPOEA, affordance_Fang18}. 
Other works tackled tasks related to EHOI detection proposing hand-centric methods \cite{Cai2016UnderstandingHM, Lending_Hand_Bambach_15, Hands_in_contact_Shan20}. 
Despite these related works have considered human-object interaction from an egocentric point of view, the EHOI detection task has not yet been defined or studied systematically in past works.
With this paper we aim at providing a definition of the task, a suitable benchmark dataset, as well as an initial evaluation of baseline approaches.

\begin{figure}[t]
\centering
\includegraphics[width=0.5\textwidth]{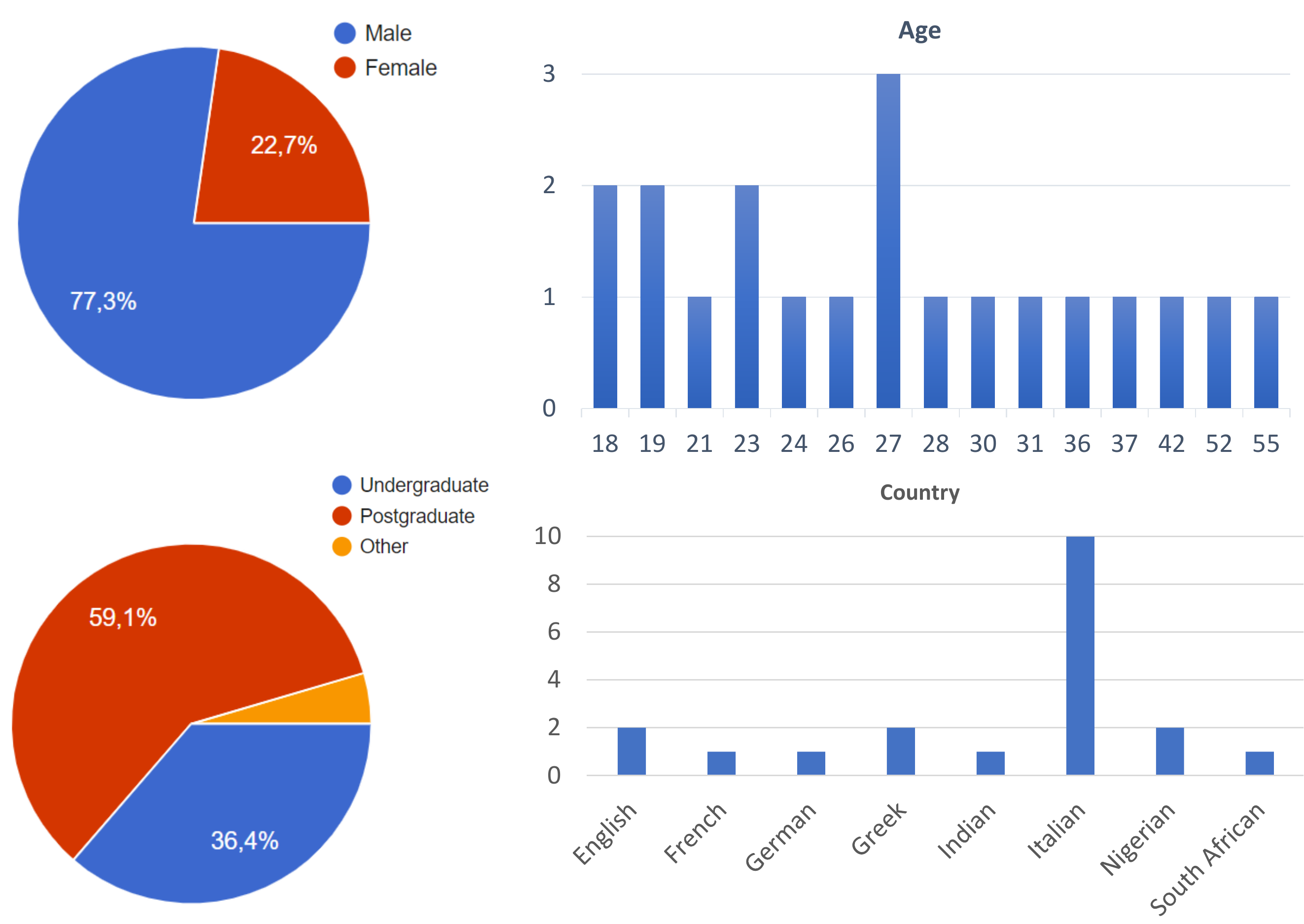}
\caption{Statistics of the 20 participants.}
\label{fig:participants}
\end{figure}
\section{MECCANO Dataset}
\label{sec:dataset}
\subsection{Data Collection}
The MECCANO dataset has been acquired in an industrial-like scenario in which subjects built a toy model of a motorbike (see Figure~\ref{fig:toy_model}). The motorbike is composed of 49 components with different shapes and sizes belonging to 19 classes. In our settings, the components \textit{A054} and \textit{A051} of Figure~\ref{fig:toy_model} have been grouped under the ``screw'' class, whereas \textit{A053}, \textit{A057} and \textit{A077} have been grouped under the ``washers'' class. As a result, we have 16 component classes\footnote{See supplementary material for more details.}. Note that multiple instances of each class are necessary to build the model. In addition, 2 tools, a \textit{screwdriver} and a \textit{wrench}, are available to facilitate the assembly of the toy model. The subjects can use the instruction booklet to understand how to build the toy model following the sequential instructions.

For the data collection, the $49$ components related to the considered $16$ classes, the $2$ tools and the instruction booklet have been placed on a table to simulate an industrial-like environment. Objects of the same component class have been grouped and placed in a heap, and heaps have been placed randomly (see Figure~\ref{fig:dataset}). Other objects not related to the toy model were present in the scene (i.e., clutter background). We have considered two types of table: a light-colored table and a dark one. The dataset has been acquired by 20 different subjects in 2 countries (Italy and United Kingdom) between May 2019 and January 2020. Participants were from $8$ different nationalities with ages between $18$ and $55$. Figure~\ref{fig:participants} reports some statistics about participants. We asked participants to sit and build the model of the motorbike. No other particular instruction was given to the participants, who were free to use all the objects placed in the table as well as the instruction booklet. Some examples of the captured data are reported in Figure~\ref{fig:dataset}.

\begin{figure}[t]
\centering
\includegraphics[width=0.45\textwidth]{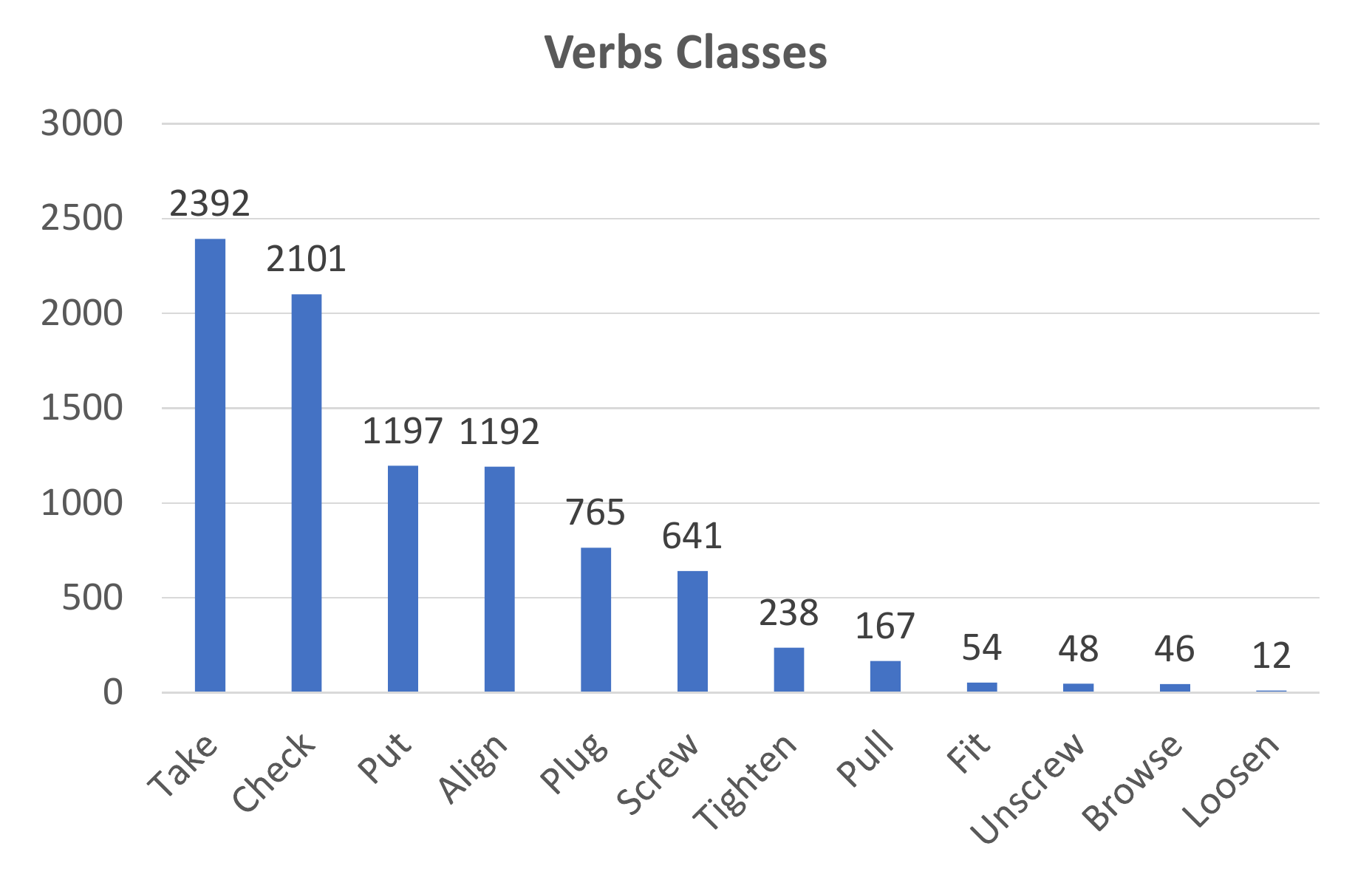}
\caption{Long-tail distribution of verbs classes.}
\label{fig:stats}
\end{figure}

\begin{figure*}[t]
\centering
\includegraphics[width=\textwidth]{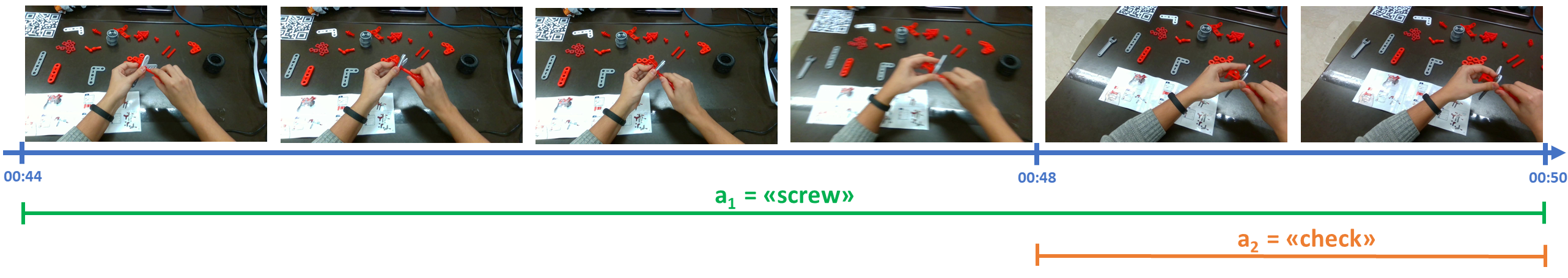}
\caption{Example of two overlapping temporal annotations along with the associated verbs.}
\label{fig:temporal}
\end{figure*}

\begin{figure*}[t]
\centering
\includegraphics[width=0.9\textwidth]{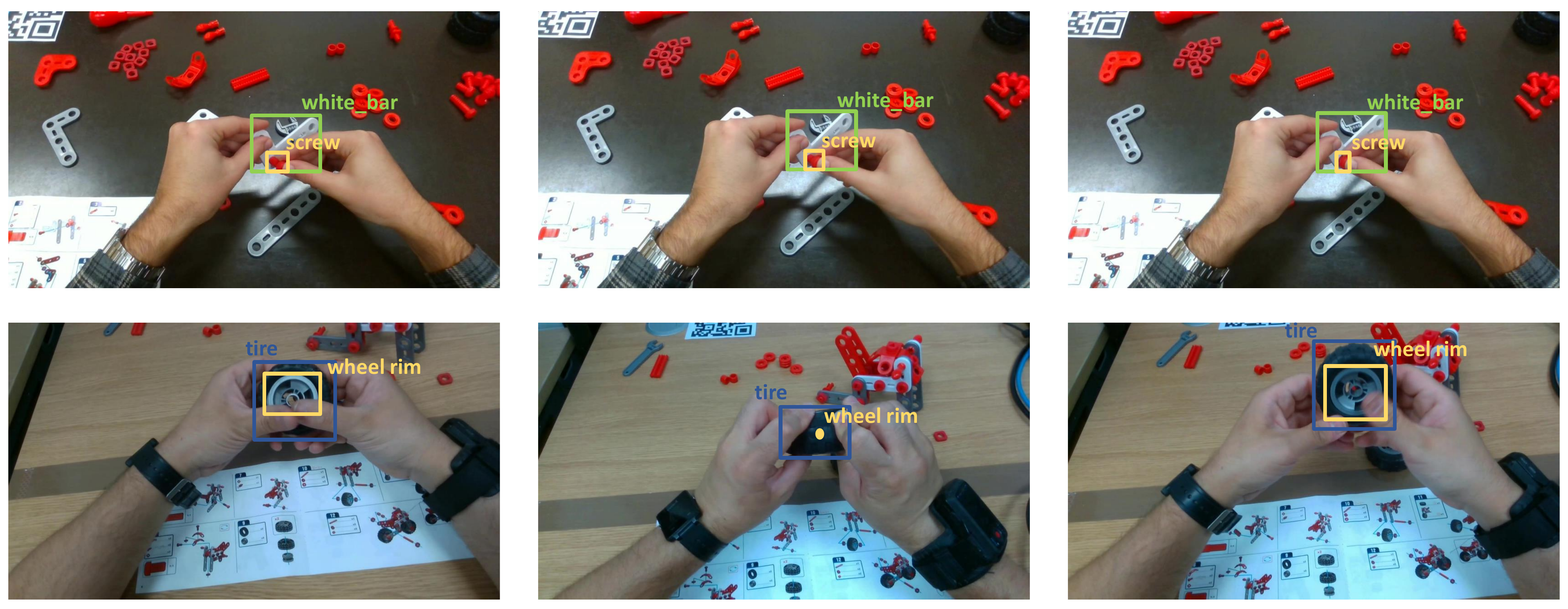}
\caption{Example of bounding box annotations for \textit{active} objects (first row) and occluded \textit{active} objects (second row).}
\label{fig:bbox}
\end{figure*}

Data was captured using an Intel RealSense SR300 device which has been mounted on the head of the participant with an headset. The headset was adjusted to control the point of view of the camera with respect to the different heights and postures of the participants, in order to have the hands located approximately in the middle of the scene to be acquired. Videos were recorded at a resolution of 1280x720 pixels and with a framerate of 12\textit{fps}. Each video corresponds to a complete assembly of the toy model starting from the 49 pieces placed on the table. The average duration of the captured videos is 21.14\textit{min}, with the longest one being 35.45\textit{min} and the shortest one being~9.23\textit{min}. \\

\subsection{Data Annotation}
We annotated the MECCANO dataset in two stages. In the first stage, we temporally annotated the occurrences of all human-object interactions indicating their start and end times, as well as a verb describing the interaction. In the second stage, we annotated the \textit{active objects} with bounding boxes for each temporal segment.

\textbf{Stage 1: Temporal Annotations \hspace{1mm}} 
We considered 12 different verbs which describe the actions performed by the participants: \textit{take, put, check, browse, plug, pull, align, screw, unscrew, tighten, loosen} and \textit{fit}. As shown in Figure~\ref{fig:stats}, the distribution of verb classes of the labeled samples in our dataset follows a long-tail distribution, which suggests that the taxonomy captures the complexity of the considered scenario.
Since a participant can perform multiple actions simultaneously, we allowed the annotated segments to overlap (see Figure~\ref{fig:temporal}). In particular, in the MECCANO dataset there are 1401 segments (15.82 \%) which overlap with at least another segment. We consider the start time of a segment as the timestamp in which the hand touches an object, changing its state from \textit{passive} to \textit{active}. The only exception is for the verb \textit{check}, in which case the user doesn't need to touch an object to perform an interaction. In this case, we annotated the start time when it is obvious from the video sequence that the user is looking at the object (see Figure~\ref{fig:temporal}). With this procedure, we annotated $8857$ video segments.

\textbf{Stage 2: Active Object Bounding Box Annotations \hspace{1mm}} 
We considered $20$ object classes which include the $16$ classes categorizing the $49$ components, the two tools (\textit{screwdriver} and \textit{wrench}), the instructions booklet and a \textit{partial\_model} class. The latter object class represents assembled components of the toy model which are not yet complete (e.g., a \textit{screw} and a \textit{bolt} fixed on a \textit{bar} which have not yet been assembled with the rest of the model\footnote{See the supplementary material for examples of partial model.}).
For each temporal segment, we annotated the \textit{active} objects in frames sampled every $0.2$ seconds. Each active object annotation consists in a \textit{(class, x, y, w, h)} tuple, where \textit{class} represents the class of the object and \textit{(x, y, w, h)} defines a bounding box around the object. 
We annotated multiple objects when they were \textit{active} simultaneously (see Figure~\ref{fig:bbox} - first row). Moreover, if an active object is occluded, even just in a few frames, we annotated it  with a \textit{(class, x, y)} tuple, specifying the class of the object and its estimated 2D position. An example of occluded active object annotation is reported in the second row of Figure~\ref{fig:bbox}. 
With this procedure, we labeled a total of 64349 frames.

\begin{table*}[]
\centering
\resizebox{0.9\textwidth}{!}{%
\begin{tabular}{l|ccccccc}
\multicolumn{1}{c|}{\textbf{Split}} & \textbf{\#Videos} & \textbf{Duration (min)} & \textbf{\%} & \textbf{\#EHOIs Segments} & \textbf{Bounding Boxes} & \textbf{Country (U.K/Italy)} & \textbf{Table (Light/Dark)} \\ \hline
Train & 11 & 236.47 & 55\% & 5057 & 37386 & 6/5 & 6/5 \\ 
Val & 2 & 46.57 & 10\% & 977 & 6983 & 1/1 & 1/1 \\ 
Test & 7 & 134.93 & 35\% & 2824 & 19980 & 4/3 & 4/3 \\ \hline
\end{tabular}%
}
\caption{Statistics of the three splits: Train, Validation and Test.}
\label{tab:splits}
\end{table*}

\begin{figure*}[t]
\centering
\includegraphics[width=\textwidth]{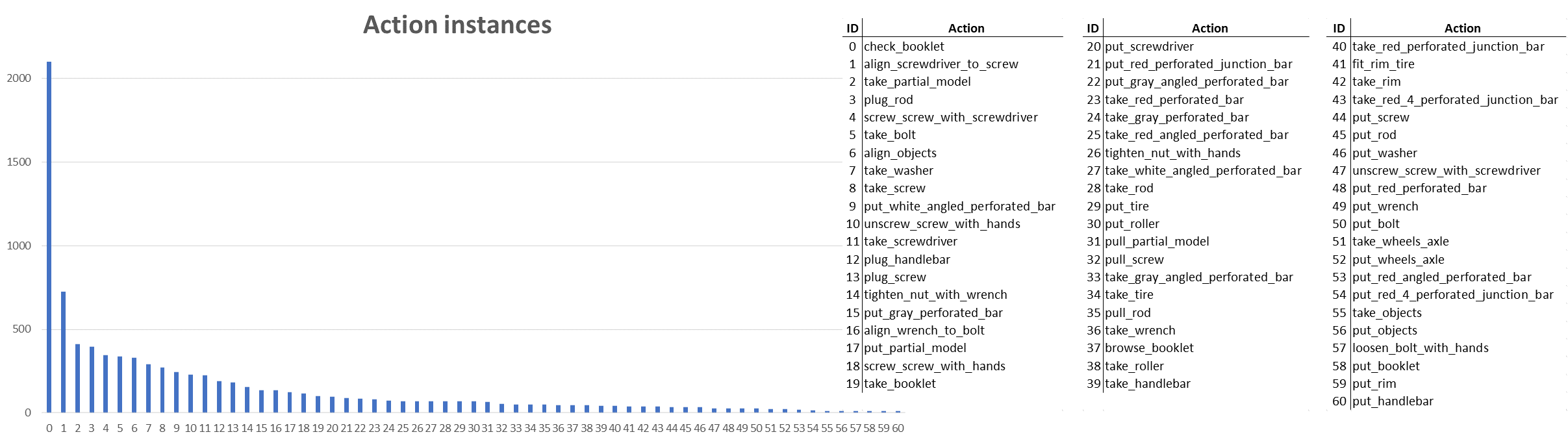}
\caption{Distribution of action instances in the MECCANO dataset.}
\label{fig:action_stats}
\end{figure*}

\textbf{Action Annotations \hspace{1mm}} 
Starting from the temporal annotations, we defined $61$ action classes\footnote{See the supplementary material for details on action class selection.}. Each action is composed by a verb and one or more objects, for example \textit{``align screwdriver to screw''} in which the verb is \textit{align} and the objects are \textit{screwdriver} and \textit{screw}. Depending on the verb and objects involved in the interaction, each temporal segment has been associated to one of the $61$ considered action classes. 
Figure~\ref{fig:action_stats} shows the list of the $61$ action classes, which follow a long-tail distribution.

\textbf{EHOI Annotations \hspace{1mm}} 
Let $O = \{o_1, o_2, ..., o_n\}$ and $V = \{v_1, v_2, ..., v_m\}$ be the sets of objects and verbs respectively. We define an Egocentric Human-Object Interaction $e$ as:

  \begin{equation} \label{eq:1}
    e = (v_h, \{o_1, o_2, ..., o_i\})
    \end{equation} 

where \begin{math}v_h \in V\end{math} is the verb characterizing the interaction and \begin{math}(o_1, o_2, ..., o_i) \subseteq O \end{math} represent the active objects involved in the interaction.
Given the previous definition, we considered all the observed combinations of verbs and objects to represent EHOIs performed by the participants during the acquisition (see examples in Figure~\ref{fig:concept}-bottom). 
Each EHOI annotation is hence composed of a verb annotation and the \textit{active} object bounding boxes. The MECCANO dataset is the first dataset of egocentric videos explicitly annotated for the EHOI detection task.

\begin{table*}[]
\centering
\resizebox{0.8\textwidth}{!}{%
\begin{tabular}{ll|l|l|l|l}
\textbf{} & \multicolumn{1}{c|}{\textbf{Top-1 Accuracy}} & \multicolumn{1}{c|}{\textbf{Top-5 Accuracy}} & \multicolumn{1}{c|}{\textbf{Avg Class Precision}} & \multicolumn{1}{c|}{\textbf{Avg Class Recall}} & \multicolumn{1}{c}{\textbf{Avg Class  F\_1-score}} \\ \hline
\multicolumn{1}{l|}{C2D \cite{fan2020pyslowfast}} & \multicolumn{1}{c|}{41.92} & \multicolumn{1}{c|}{71.95} & \multicolumn{1}{c|}{37.6} & \multicolumn{1}{c|}{38.76} & \multicolumn{1}{c}{36.49}  \\
\multicolumn{1}{l|}{I3D \cite{Carreira2017QuoVA}} & \multicolumn{1}{c|}{42.51} & \multicolumn{1}{c|}{72.35} & \multicolumn{1}{c|}{40.04} & \multicolumn{1}{c|}{40.42} & \multicolumn{1}{c}{38.88}  \\
\multicolumn{1}{l|}{SlowFast \cite{feichtenhofer2018slowfast}} & \multicolumn{1}{c|}{\textbf{42.85}} & \multicolumn{1}{c|}{\textbf{72.47}} & \multicolumn{1}{c|}{\textbf{42.11}} & \multicolumn{1}{c|}{\textbf{41.48}} & \multicolumn{1}{c}{\textbf{41.05}} \\ \hline
\end{tabular}%
}
\caption{Baseline results for the action recognition task.}
\label{tab:action}
\end{table*}

\section{Benchmarks and Baseline Results}
\label{sec:benchmarks}
The MECCANO dataset is suitable to study a variety of tasks, considering the challenging industrial-like scenario in which it was acquired. In this paper, we consider four tasks for which we provide baseline results: 1) \textit{Action Recognition}, 2) \textit{Active Object Detection}, 3) \textit{Active Object Recognition} and 4) \textit{Egocentric Human-Object Interaction (EHOI) Detection}. While some of these tasks have been considered in previous works, none of them has been studied in industrial scenarios from the egocentric perspective. Moreover, it is worth noting that the EHOI Detection task has never been treated in previous works.
We split the dataset into three subsets (\textit{Training, Validation} and \textit{Test}) designed to balance the different types of desks (light, dark) and countries in which the videos have been acquired (IT, U.K.). Table~\ref{tab:splits} reports some statistics about the three splits, such as the number of videos, the total duration (in seconds), the number of temporally annotated EHOIs and the number of bounding box annotations.

\subsection {Action Recognition}
\textbf{Task:} Action Recognition consists in determining the action performed by the camera wearer from an egocentric video segment. Specifically, let \begin{math}C_a = \{c_1, c_2, ..., c_n\} \end{math} be the set of action classes and let \begin{math}A_i = [t_{s_i}, t_{e_i}]\end{math} be a video segment,  where $t_{s_i}$ and $t_{e_i}$ are the start and the end times of the action respectively. The aim is to assign the correct action class $c_i \in C_a$ to the segment $A_i$.\\ 
\textbf{Evaluation Measures:} We evaluate action recognition using Top-1 and Top-5 accuracy computed on the whole test set. As class-aware measures, we report class-mean precision, recall and $F_1$-score.  \\
\textbf{Baselines:} We considered 2D CNNs as implemented in the PySlowFast library \cite{fan2020pyslowfast} (C2D), I3D \cite{Carreira2017QuoVA} and SlowFast \cite{feichtenhofer2018slowfast} networks, which are state-of-the-art methods for action recognition. In particular, for all baselines we used the PySlowFast implementation based on a ResNet-50 \cite{He2015_ResNet} backbone pre-trained on Kinetics \cite{Kinetics_2017}. See supplementary material for implementation details. \\
\textbf{Results:} Table~\ref{tab:action} reports the results obtained by the baselines for the action recognition task. All baselines obtained similar performance in terms of Top-1 and Top-5 accuracy with SlowFast networks achieving slightly better performance.
Interestingly, performance gaps are more consistent when we consider precision, recall and $F_1$ scores, which is particularly relevant given the long-tailed distribution of actions in the proposed dataset (see Figure~\ref{fig:action_stats}). Note that, in our benchmark, SlowFast obtained the best results with a Top-1 accuracy of 47.82 and an $F_1$-score of 41.05. 
See supplementary material for qualitative results.
In general, the results suggest that action recognition with the MECCANO dataset is challenging and offers a new scenario to compare action recognition algorithms.

\begin{table}[]
\centering
\resizebox{0.5\textwidth}{!}{%
\begin{tabular}{l|c}
\multicolumn{1}{c|}{\textbf{Method}} & \multicolumn{1}{l}{\textbf{AP (IoU \textgreater 0.5)}} \\ \hline
Hand Object Detector \cite{Hands_in_contact_Shan20} & 11.17\% \\
Hand Object Detector \cite{Hands_in_contact_Shan20} (Avg dist.)  & 11.10\% \\
Hand Object Detector \cite{Hands_in_contact_Shan20} (All dist) & 11.34\% \\
Hand Object Detector \cite{Hands_in_contact_Shan20} + Objs re-training  & 20.18\% \\
Hand Object Detector \cite{Hands_in_contact_Shan20} + Objs re-training (Avg dist.)  & 33.33\% \\
Hand Object Detector \cite{Hands_in_contact_Shan20} + Objs re-training (All dist.)  & \textbf{38.14\%} \\ \hline
\end{tabular}%
}
\caption{Baseline results for the \textit{active} object detection task.}
\label{tab:active_det}
\end{table}


\subsection {Active Object Detection}
\textbf{Task:} The aim of the Active Object Detection task is to detect all the \textit{active} objects involved in EHOIs. 
Let \begin{math} O_{act} = \{o_1, o_2, ..., o_n\} \end{math} be the set of \textit{active} objects in the image. The goal is to detect with a bounding box each \textit{active} object $o_i \in O_{act}$. \\
\textbf{Evaluation Measures:} As evaluation measure, we use Average Precision~(AP), which is used in standard object detection benchmarks. We set the IoU threshold equal to~$0.5$ in our experiments. \\
\textbf{Baseline:} We considered the Hand-Object Detector proposed in \cite{Hands_in_contact_Shan20}. The model has been designed to detect hands and objects when they are in contact. This architecture is based on Faster-RCNN \cite{ren2015faster} and predicts a box around the visible human hands, as well as boxes around the objects the hands are in contact with and a link between them. We used the Hand-Object Detector \cite{Hands_in_contact_Shan20} pretrained on EPIC-Kitchens \cite{Damen2018EPICKITCHENS}, EGTEA \cite{Li2018_EGTEA-GAZE+} and CharadesEGO \cite{Sigurdsson2018Charades} as provided by authors \cite{Hands_in_contact_Shan20}. The model has been trained to recognize hands and to detect the \textit{active} objects regardless their class. Hence, it should generalize to others domains.
With default parameters, the Hand-Object Detector can find at most two \textit{active} objects in contact with hands. Since our dataset tends to contain more \textit{active} objects in a single EHOI (up to 7), we consider two variants of this model by changing the threshold on the distance between hands and detected objects. In the first variant, the threshold is set to the average distance between hands and \textit{active} objects on the MECCANO dataset. We named this variant ``\textit{Avg distance}''. In the second variant, we removed the thresholding operation and considered all detected objects as \textit{active} objects. We named this variant ``\textit{All objects}''. 
We further adapted the Hand-Object Detector \cite{Hands_in_contact_Shan20} re-training the Faster-RCNN component to detect all \textit{active} objects of the MECCANO dataset. See supplementary material for implementation details.\\
\textbf{Results:} Table~\ref{tab:active_det} shows the results obtained by the \textit{active} object detection task baselines. The results highlight that the Hand-Object Detector \cite{Hands_in_contact_Shan20} is not able to generalize to a domain different than the one on which it was trained. All the three variants of the Hand-Object Detector using the original object detector obtained an AP approximately equal to 11\% (first three rows of Table~\ref{tab:active_det}). Re-training the object detector on the MECCANO dataset allowed to improve performance by significant margins. In particular, using the standard distance threshold value, we obtained an AP of 20.18\%. If we consider the average distance as the threshold to discriminate \textit{active} and \textit{passive} objects, we obtain an AP of 33.33\%. Removing the distance threshold (last row of Table~\ref{tab:active_det}), allows to outperform all the previous results obtaining an AP equal to 38.14\%. This suggests that adapting the general object detector to the challenging domain of the proposed dataset is key to performance. Indeed, training the object detector to detect only \textit{active} objects in the scene already allows to obtain reasonable results, while there still space for improvement.

\subsection {Active Object Recognition}
\textbf{Task:} The task consists in detecting and recognizing the \textit{active} objects involved in EHOIs considering the $20$ object classes of the MECCANO dataset.
Formally, let \begin{math} O_{act} = \{o_1, o_2, ..., o_n\}\end{math} be the set of \textit{active} objects in the image and let \begin{math} C_{o} = \{c_1, c_2, ..., c_m\} \end{math} be the set of object classes. The task consists in detecting objects $o_i \in O_{act}$ and assigning them the correct class label $c \in C_{o}$. \\
\textbf{Evaluation Measures:} We use mAP \cite{PascalVOC_Zisserman_15} with threshold on IoU equal to $0.5$ for the evaluations.\\
\textbf{Baseline:} As a baseline, we used a standard Faster-RCNN \cite{ren2015faster} object detector. For each image the object detector predicts \textit{(x, y, w, h, class)} tuples which represent the object bounding boxes and the associated classes. See supplementary material for implementation details. \\
\textbf{Results:} Table~\ref{tab:active_rec} reports the results obtained with the baseline in the \textit{Active} Object Recognition task.
We report the AP values for each class considering all the videos belonging to the test set of the MECCANO dataset. The last column shows the average of the AP values for each class and the last row reports the mAP value for the test set. The mAP was computed as the average of the mAP values obtained in each test video. AP values in the last column show that large objects are easier to recognize (e.g. \textit{instruction booklet: 46.48\%; screwdriver: 60.50\%; tire: 58.91\%; rim: 50.35\%}). Performance suggests that the proposed dataset is challenging due to the presence of small objects.
We leave the investigation of more specific approaches to active object detection to future studies.

\begin{table}[]
\centering
\resizebox{0.8\columnwidth}{!}{%
\begin{tabular}{clc}
\multicolumn{1}{l|}{\textbf{ID}} & \multicolumn{1}{c|}{\textbf{Class}} & \textbf{AP (per class)} \\ \hline
\multicolumn{1}{c|}{0} & \multicolumn{1}{l|}{instruction booklet} & 46.18\% \\
\multicolumn{1}{c|}{1} & \multicolumn{1}{l|}{gray\_angled\_perforated\_bar} & 09.79\% \\
\multicolumn{1}{c|}{2} & \multicolumn{1}{l|}{partial\_model} & 36.40\% \\
\multicolumn{1}{c|}{3} & \multicolumn{1}{l|}{white\_angled\_perforated\_bar} & 30.48\% \\
\multicolumn{1}{c|}{4} & \multicolumn{1}{l|}{wrench} & 10.77\% \\
\multicolumn{1}{c|}{5} & \multicolumn{1}{l|}{screwdriver} & 60.50\% \\
\multicolumn{1}{c|}{6} & \multicolumn{1}{l|}{gray\_perforated\_bar} & 30.83\% \\
\multicolumn{1}{c|}{7} & \multicolumn{1}{l|}{wheels\_axle} & 10.86\% \\
\multicolumn{1}{c|}{8} & \multicolumn{1}{l|}{red\_angled\_perforated\_bar} & 07.57\% \\
\multicolumn{1}{c|}{9} & \multicolumn{1}{l|}{red\_perforated\_bar} & 22.74\% \\
\multicolumn{1}{c|}{10} & \multicolumn{1}{l|}{rod} & 15.98\% \\
\multicolumn{1}{c|}{11} & \multicolumn{1}{l|}{handlebar} & 32.67\% \\
\multicolumn{1}{c|}{12} & \multicolumn{1}{l|}{screw} & 38.96\% \\
\multicolumn{1}{c|}{13} & \multicolumn{1}{l|}{tire} & 58.91\% \\
\multicolumn{1}{c|}{14} & \multicolumn{1}{l|}{rim} & 50.35\% \\
\multicolumn{1}{c|}{15} & \multicolumn{1}{l|}{washer} & 30.92\% \\
\multicolumn{1}{c|}{16} & \multicolumn{1}{l|}{red\_perforated\_junction\_bar} & 19.80\% \\
\multicolumn{1}{c|}{17} & \multicolumn{1}{l|}{red\_4\_perforated\_junction\_bar} & 40.82\% \\
\multicolumn{1}{c|}{18} & \multicolumn{1}{l|}{bolt} & 23.44\% \\
\multicolumn{1}{c|}{19} & \multicolumn{1}{l|}{roller} & 16.02\% \\ \hline
\multicolumn{1}{l}{} &  & \multicolumn{1}{l}{} \\ \cline{2-3} 
\multicolumn{1}{l}{} & \multicolumn{1}{c|}{\textbf{mAP}} & 30.39\%
\end{tabular}%
}
\caption{Baseline results for the \textit{active} object recognition task.}
\label{tab:active_rec}
\end{table}

\subsection {EHOI Detection}
\begin{figure}[t]
	\centering
	\includegraphics[width=0.5\textwidth]{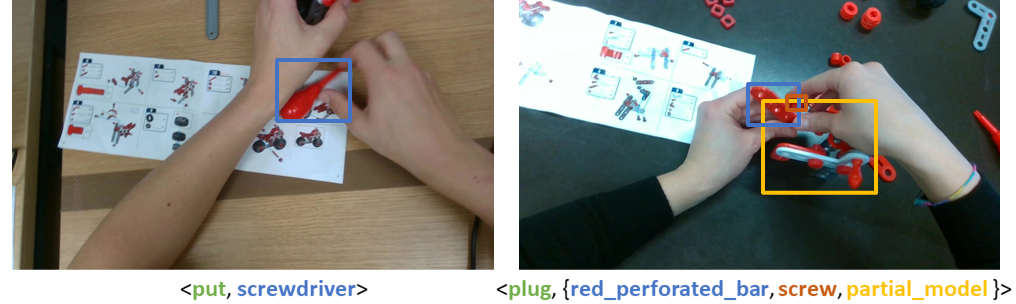}
	\caption{Qualitative results for the EHOI detection task.}
	\label{fig:EHOI_qual}
\end{figure}

\textbf{Task:} The goal is to determine egocentric human-object interactions (EHOI) in each image. Given the definition of EHOIs as $<$verb, objects$>$ pairs (see Equation~\ref{eq:1}), methods should  detect and recognize all the \textit{active} objects in the scene, as well as the verb describing the action performed by the human. \\
\textbf{Evaluation Measures:} 
Following \cite{Gupta2015VisualSR, Gkioxari2018DetectingAR}, we use the \textit{``role AP''} as an evaluation measure. 
Formally, a detected EHOI is considered as a true positive if 1) the predicted object bounding box has a IoU of 0.5 or higher with respect to a ground truth annotation and 2) the predicted verb matches with the ground truth. 
Note that only the \textit{active} object bounding box location (not the correct class) is considered in this measure. 
Moreover, we used different values of IoU (e.g., 0.5, 0.3 and 0.1) to compute the \textit{``role AP''}.\\
\textbf{Baseline:} We adopted three baselines for the EHOI detection task. 
The first one is based on InteractNet \cite{Gkioxari2018DetectingAR}, which is composed by three branches: 1) the ``human-branch'' to detect the humans in the scene, 2) the ``object-branch'' to detect the objects and 3) the ``interaction-branch' which predicts the verb of the interaction focusing on the humans and objects appearance. The second one is an extension of InteractNet which also uses context features derived from the whole input frame to help the ``interaction-branch'' in verb prediction. The last baseline is based on the combination of a SlowFast network \cite{fan2020pyslowfast} trained to predict the verb of the EHOI considering the spatial and temporal dimensions, and Faster-RCNN \cite{ren2015faster} which detects and recognizes all \textit{active} objects in the frame. 
See supplementary material for implementation details.\\
\textbf{Results:} Table~\ref{tab:EHOI_det} reports the results obtained by the baselines on the test set for the EHOI detection task. The InteractNet method obtains low performance on this task with a mAP role of 4.92\%. Its extension with context features, slightly improves the performance with a mAP role of 8.45\%, whereas SlowFast network with Faster-RCNN achieved best results with a mAP equal to 25.93\%. 
The results highlight that current state-of-the-art approaches developed for the analysis of still images in third person scenarios are unable to detect EHOIs in the proposed dataset, which is likely due to the presence of multiple tiny objects involved simultaneously in the EHOI and to the actions performed.
On the contrary, adding the ability to process video clips with SlowFast allows for significant performance boosts.
Figure~\ref{fig:EHOI_qual} shows qualitative results obtained with the SlowFast+Faster-RCNN baseline. Note that in the second example the method correctly predicted all the objects involved simultaneously in the EHOI. 
Despite promising performance of the suggested baseline, the proposed EHOI detection task needs more investigation due to the challenging nature of the considered industrial-like domain.

\begin{table}[]
\centering
\resizebox{0.5\textwidth}{!}{%
\begin{tabular}{l|lll}
 & \multicolumn{3}{c|}{\textbf{mAP role}} \\ \hline
\textbf{Model} & \textbf{IoU $\geq$ 0.5} & \textbf{IoU $\geq$ 0.3} & \textbf{IoU $\geq$ 0.1} \\ \hline
InteractNet \cite{Gkioxari2018DetectingAR} & \multicolumn{1}{c}{04.92\%} & \multicolumn{1}{c}{05.30\%} & \multicolumn{1}{c}{05.72\%} \\
InteractNet \cite{Gkioxari2018DetectingAR} + Context & \multicolumn{1}{c}{08.45\%} & \multicolumn{1}{c}{09.01\%} & \multicolumn{1}{c}{09.45\%} \\ 
SlowFast \cite{feichtenhofer2018slowfast} + Faster-RCNN \cite{ren2015faster} & \multicolumn{1}{c}{\textbf{25.93\%}} & \multicolumn{1}{c}{\textbf{28.04\%}} & \multicolumn{1}{c}{\textbf{29.65\%}} \\\hline
\end{tabular}%
}
\caption{Baseline results for the EHOI detection task.}
\label{tab:EHOI_det}
\end{table}

\section{Conclusion}
\label{sec:conclusion}
We proposed MECCANO, the first dataset to study egocentric human-object interactions (EHOIs) in an industrial-like scenario. 
We publicly release the dataset with both temporal (action segments) and spatial (active object bounding boxes) annotations considering a taxonomy of $12$ verbs, $20$ nouns and $61$ unique actions.
In addition, we defined the Egocentric Human-Object Interaction (EHOI) detection task and performed baseline experiments to show the potential of the proposed dataset on four challenging tasks: action recognition, \textit{active} object detection, \textit{active} object recognition and EHOI detection.
Future works will explore approaches for improved performance on this challenging data.

\section*{Acknowledgments}
This research has been supported by MIUR PON PON R\&I 2014-2020 - Dottorati innovativi con caratterizzazione industriale, by MIUR AIM - Attrazione e Mobilita Internazionale Linea 1 - AIM1893589 - CUP: E64118002540007, and by MISE - PON I\&C 2014-2020 - Progetto ENIGMA  - Prog n. F/190050/02/X44 – CUP: B61B19000520008.

\section*{\uppercase{Supplementary Material}}
\label{sec:supp_material}
This document is intended for the convenience of the reader and reports additional information about the proposed dataset, the annotation stage, as well as implementation details related to the performed experiments. This supplementary material is related to the following submission:
\begin{itemize}
    \item F. Ragusa, A. Furnari, S. Livatino, G. M. Farinella. The MECCANO Dataset: Understanding Human-Object Interactions from Egocentric Videos in an Industrial-like Domain, submitted to IEEE Winter Conference on Applications of Computer Vision (WACV), 2021.  
\end{itemize}

The remainder of this document is organized as follows. Section~\ref{ref:dataset} reports additional details about data collection and annotation. Section~\ref{ref:implementation} provides implementation details of the compared methods. Section~\ref{ref:qualitative} reports additional qualitative results.
\section{Additional details on the MECCANO Dataset}
\label{ref:dataset}
\subsection{Component classes and grouping}
\begin{figure}[t]
\centering
\includegraphics[width=0.8\columnwidth]{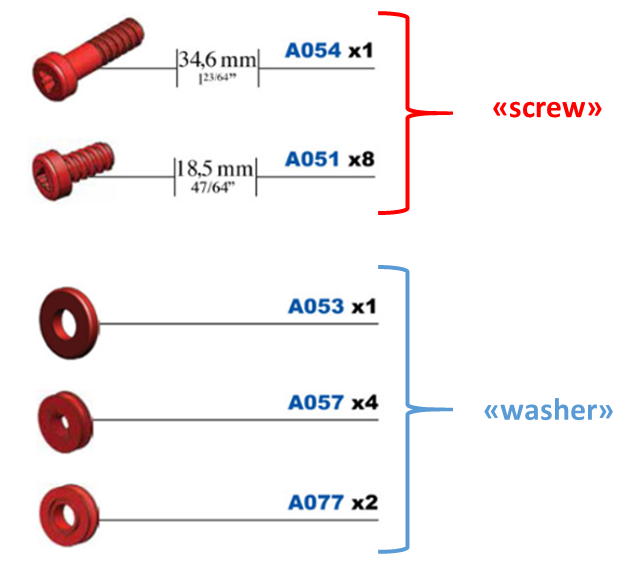}
\caption{Grouped pieces belonging to \textit{screw} and \textit{washer} classes.}
\label{fig:groups}
\end{figure}

The toy motorbike used for our data collection is composed of 49 components belonging to 19 classes (Figure~\ref{fig:toy_model}), plus two tools. 
In our settings, we have grouped two types of components which are similar in their appearance and have similar roles in the assembly process. Figure~\ref{fig:groups} illustrates the two groups. Specifically, we grouped A054 and A051 under the ``screw'' class. These two types of components only differ in their lengths. We also grouped A053, A057 and A077 under the ``washers'' class. Note that these components only differ in the radius of their holes and in their thickness. 

\begin{figure*}[t]
\centering
\includegraphics[width=0.8\textwidth]{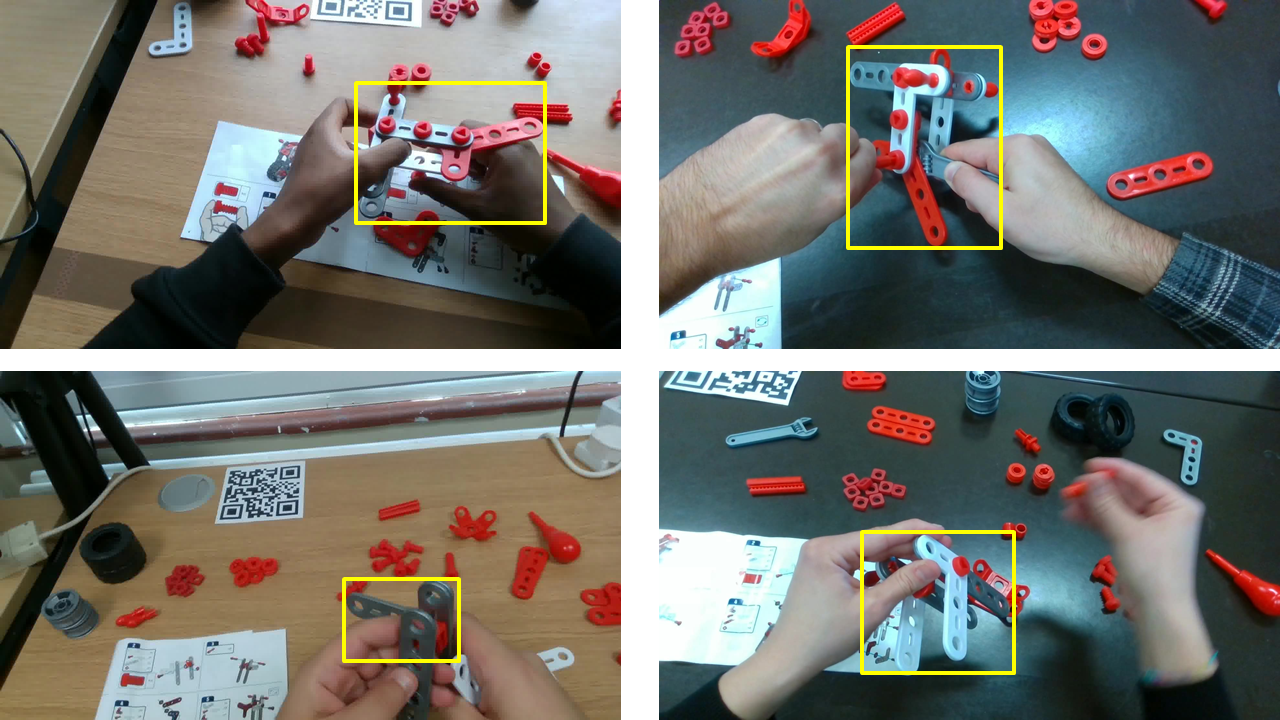}
\caption{Examples of objects belonging to the partial model class.}
\label{fig:partial}
\end{figure*}

As a results, we have 20 object classes in total: 16 classes are related to the 49 motorbike components, whereas the others are associated to the two tools, to the instruction booklet and to a partial model class, which indicates a set of components assembled together to form a part of the model (see Figure~\ref{fig:partial} ).

\subsection{Data Annotation}
\begin{figure}[t]
\centering
\includegraphics[width=\columnwidth]{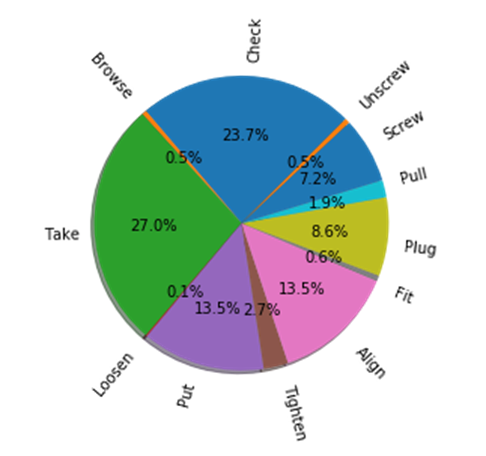}
\caption{Fractions of instances of each verb in the MECCANO dataset.}
\label{fig:verbs}
\end{figure}

\textbf{Verb Classes and Temporal Annotations \hspace{1mm}} 
We considered $12$ verb classes which describe all the observed actions performed by the participants during the acquisitions. Figure~\ref{fig:verbs} reports the percentage of the temporally annotated instances belonging to the $12$ verb classes. The considered verb classes are: \textit{take, put, check, browse, plug, pull, align, screw, unscrew, tighten, loosen} and \textit{fit}.  
We used the ELAN Annotation tool~\cite{ELAN} to annotate a temporal segment around each instance of an action. Each segment has been associated to the verb which best described the contained action.

\begin{figure*}[t]
\centering
\includegraphics[width=\textwidth]{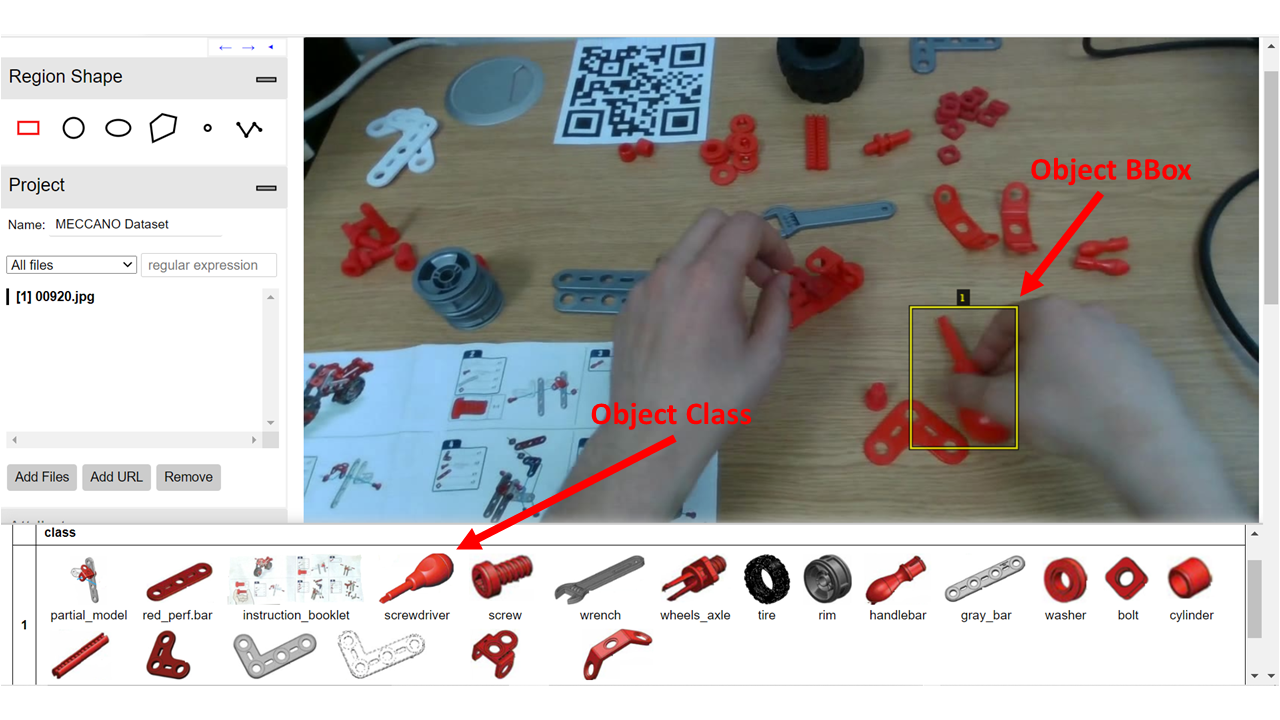}
\caption{Customized VIA project to support the labeling of active objects. Annotators were presented with a panel which allowed them to identify object classes through their thumbnails.}
\label{fig:VIA}
\end{figure*}

\begin{figure}[t]
\centering
\includegraphics[width=0.4\textwidth]{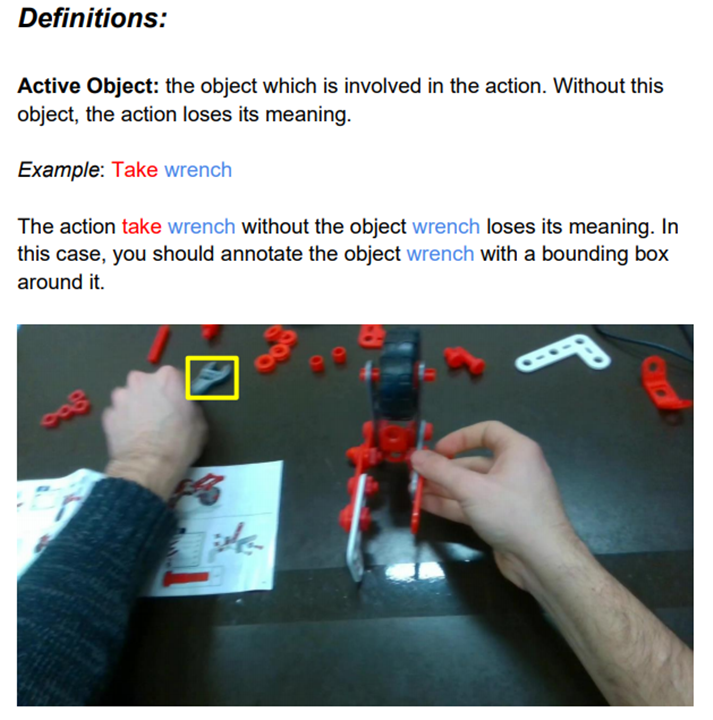}
\caption{\textit{Active} object definition given to the labelers for the \textit{active} object bounding box annotation stage.}
\label{fig:rules}
\end{figure}

\textbf{Active Object Bounding Box Annotations \hspace{1mm}}
For each annotated video segment, we sampled frames every $0.2$ seconds. 
Each of these frames has been annotated to mark the presence of all \textit{active} objects with bounding boxes and related component class label. 
To this aim, we used VGG Image Annotator (VIA) \cite{dutta2019vgg} with a customized project which allowed annotators to select component classes from a dedicated panel showing the thumbnails of each of the $20$ object classes to facilitate and speed up the selection of the correct object class. Figure~\ref{fig:VIA} reports an example of the customized VIA interface. Moreover, to support annotators and reduce ambiguities, we prepared a document containing a set of fundamental rules for the annotations of \textit{active} objects, where we reported the main definitions (e.g., active object, occluded active object, partial\_model) along with visual examples. Figure~\ref{fig:rules} reports an example of such instructions.

\begin{figure*}[t]
\centering
\includegraphics[width=\textwidth]{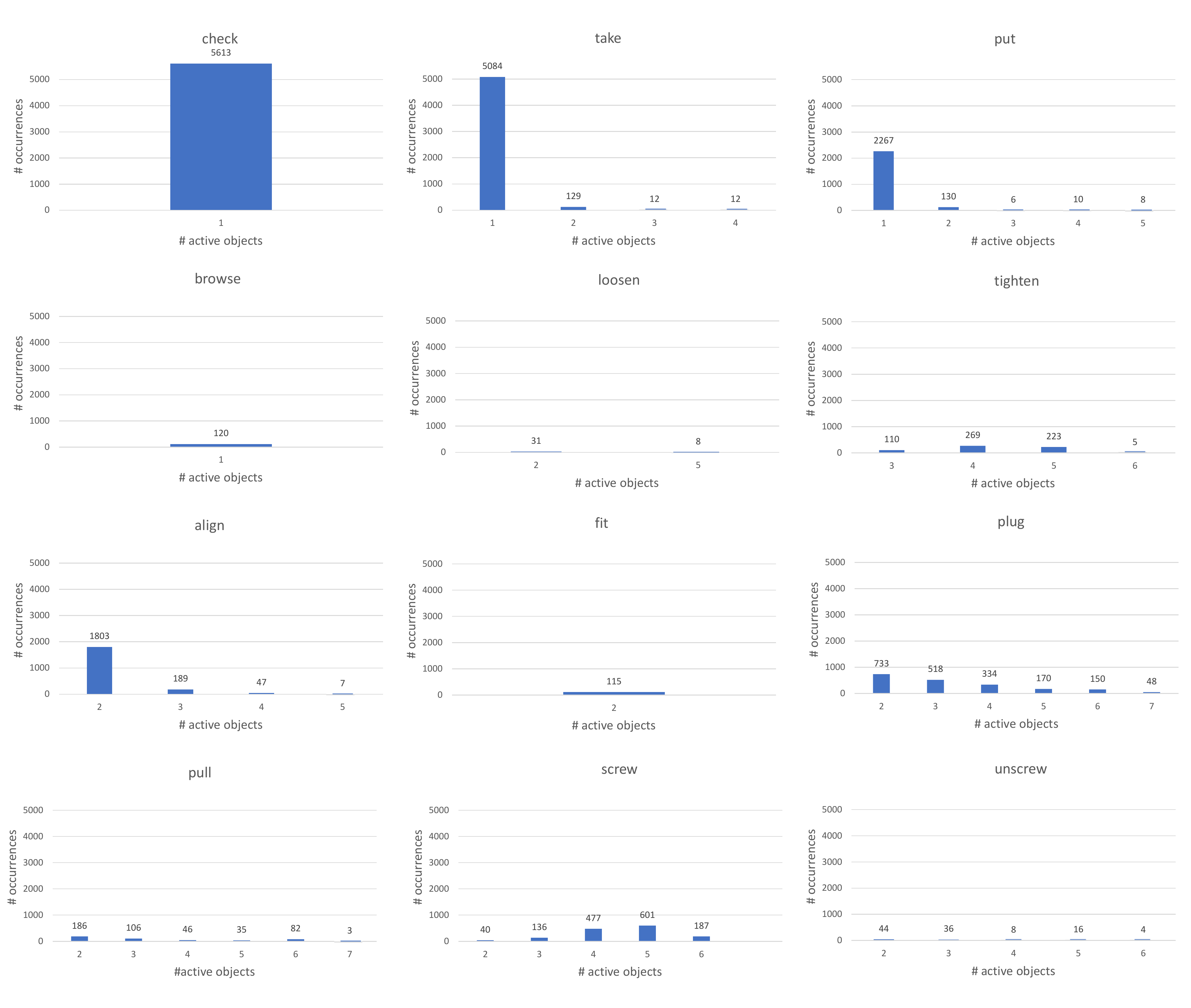}
\caption{Number of objects and occurrences of \textit{active} objects related to each verb.}
\label{fig:actions}
\end{figure*}

\textbf{Action Annotation \hspace{1mm}}
In the MECCANO dataset, an action can be seen as a combination of a verb and a set of nouns (e.g., ``take wrench''). We analyzed the combinations of our $12$ verb classes and $20$ object classes to find a compact, yet descriptive set of actions classes. 
The action class selection process has been performed in two stages. 
In the first stage, we obtained the distributions of the number of active objects generally occurring with each of the $12$ verbs. The distributions are shown in Figure~\ref{fig:actions}. For example, the dataset contains $120$ instances of ``browse'' (second row - first column), which systematically involves one single object.
Similarly, most of the instance of ``take'' appear with $1$ object, while few instances have $2-3$ objects.

\begin{figure*}[t]
\centering
\includegraphics[width=0.8\textwidth]{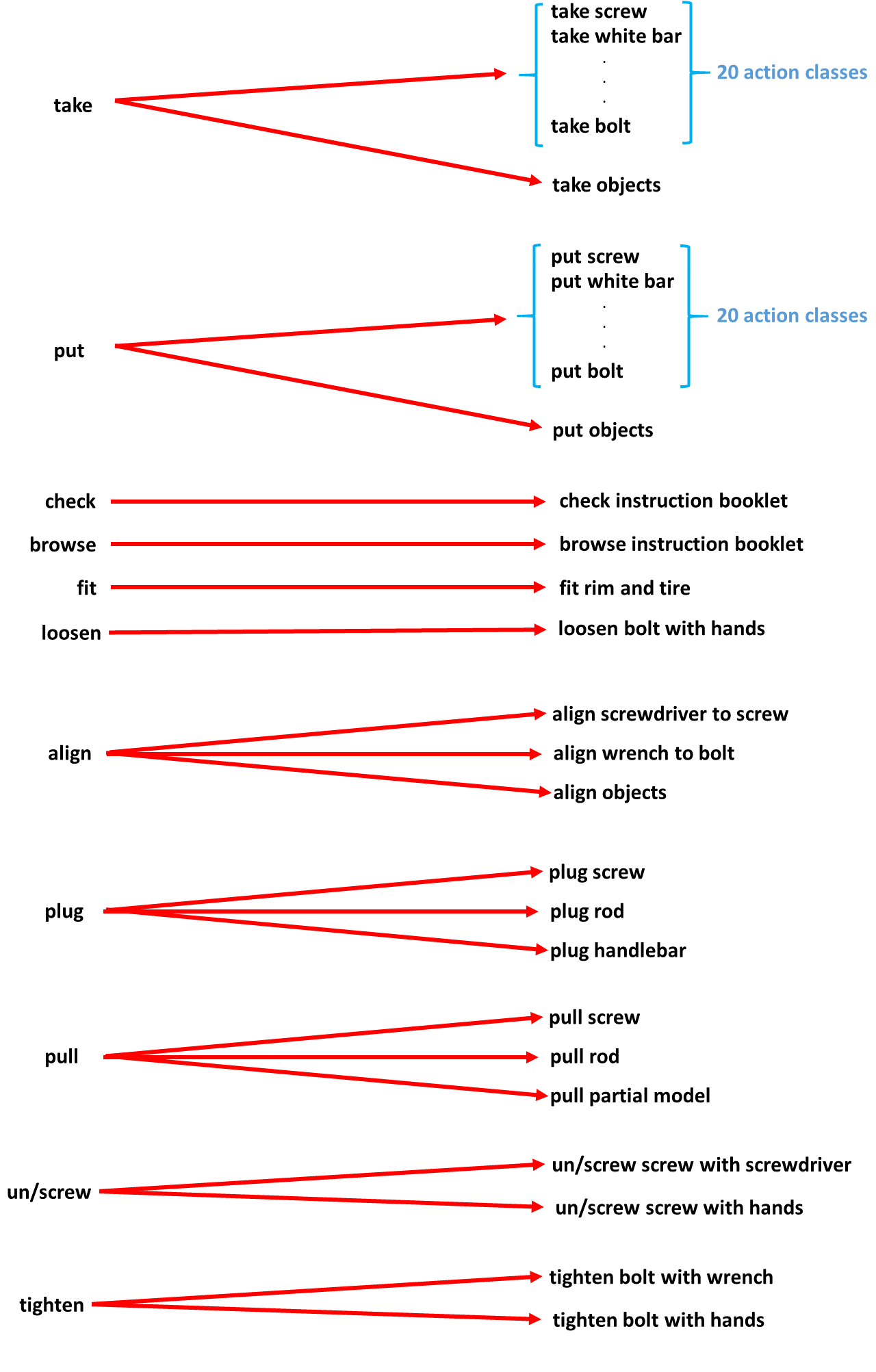}
\caption{$61$ action classes definition from the $12$ verb classes and the analysis performed observing the participant behavior.}
\label{fig:actions1}
\end{figure*}

In the second stage, we selected a subset of actions from all combinations of verbs and nouns. Figure~\ref{fig:actions1} reports all the action classes obtained from the 12 verbs classes of the MECCANO dataset as discussed in the following.
Let \begin{math} O = \{o_1, o_2, ..., o_n\} \end{math} and \begin{math} V = \{v_1, v_2, ..., v_m\} \end{math} be the set of the objects and verb classes respectively.
For each verb $v \in V$, we considered all the object classes $o \in O$ involved in one or more temporal segments labeled with verb $v$. We considered the following rules:

\begin{itemize}
    \item \textbf{Take and put}: We observed that all the objects $o \in O$ occurring with $v=take$ are taken by participants while they build the motorbike. Hence, we first defined 20 action classes as $(v, o)$ pairs (one for each of the available objects). 
    Since subjects can take more than one object at a time, we added an additional ``take objects'' action class when two or more objects are taken simultaneously. 
    The same behavior has been observed for the verb $ v = put$. Hence, we similarly defined 21 action classes related to this verb. 
    \item \textbf{Check and browse}: We observed that verbs $v = check$ and $v = browse$ always involve only the object $o = instruction$  $booklet$. Hence, we defined the two action classes \textit{check instruction booklet} and \textit{browse instruction booklet}.
    \item \textbf{Fit}: When the verb is $v = fit$, there are systematically two objects involved simultaneously (i.e., $o = rim$ and $o = tire$). Hence, we defined the action class \textit{fit rim and tire}. 
    \item \textbf{Loosen}: We observed that participants tend to loosen bolts always with the hands. We hence defined the action class \textit{loosen bolt with hands}.

\item \textbf{Align}: We observed that participants tend to align the screwdriver tool with the screw before starting to screw, as well as the wrench tool with the bolt before tightening it. Participants also tended to align objects to be assembled to each other. 
From these observations, we defined three action classes related to the verb $v = align$: \textit{align screwdriver to screw}, \textit{align wrench to bolt} and \textit{align objects}.

\item \textbf{Plug}: We found three main uses of verb $v=plug$ related to the objects $o = screw$, $o = rod$ and $o = handlebar$. Hence, we defined three action classes: \textit{plug screw}, \textit{plug rod} and \textit{plug handlebar}.

\item \textbf{Pull}: Similar observations apply to verb $v = pull$. Hence we defined three action classes involving ``pull'': \textit{pull screw}, \textit{pull rod} and \textit{pull partial model}.

\item \textbf{Screw and unscrew}: The main object involved in actions characterized by the verbs $v = screw$ and $v = unscrew$ is $o = screw$. Additionally, the screw or unscrew action can be performed with a screwdriver or with hands. Hence, we defined four action classes \textit{screw screw with screwdriver, screw screw with hands, unscrew screw with screwdriver} and \textit{unscrew screw with hands}.

\item \textbf{Tighten}: Similar observation holds for the verb $v = tighten$, the object $o = bolt$ and the tool $o = wrench$. We hence defined the following two action classes: \textit{tighten bolt with wrench} and \textit{tighten bolt with hands}.
\end{itemize}

In total, we obtained 61 action classes composing the MECCANO dataset.

\section{Baseline Implementation Details}
\label{ref:implementation}

\subsection {Action Recognition}
The goal of action recognition is to classify each action segment into one of the $61$ action classes of the MECCANO dataset.
The SlowFast, C2D and I3D baselines considered in this paper all require fixed-length clips at training time. Hence, we temporally downsample or upsample uniformly each video shot before passing it to the input layer of the network. The average number of frames in a video clip in the MECCANO dataset is $26.19$. For SlowFast network, we set $\alpha=4$ and $\beta=\frac{1}{8}$. 
We set the batch-size to $12$ for C2D and I3D, we used a batch-size of $20$ for SlowFast. 
We trained C2D, I3D and SlowFast networks on 2 NVIDIA V100 GPUs for $80$, $70$ and $40$ epochs with learning rates of 0.01, 0.1 and 0.0001 respectively. 
These settings allowed all baselines to converge. 

\subsection {Active Object Detection}
We trained Faster-RCNN on the training and validation sets using the provided \textit{active} object labels. We set the learning rate to 0.005 and trained Faster-RCNN with a ResNet-101 backbone and Feature Pyramid Network for 100K iterations on 2 NVIDIA V100 GPUs. We used the Detectron2 implementation \cite{wu2019detectron2}. The model is trained to recognize objects along with their classes. However, for the active object detection task, we ignore output class names and only consider a single ``active object'' class.

\subsection {Active Object Recognition}
We used the same model adopted for the Active Object Detection task, retaining also object classes at test time.

\begin{table*}[]
\centering
\resizebox{0.9\textwidth}{!}{%
\begin{tabular}{clcccccccc}
\multicolumn{1}{l|}{\textbf{ID}} & \multicolumn{1}{c|}{\textbf{Class\textbackslash{}Video}} & \textbf{0008} & \textbf{0009} & \textbf{0010} & \textbf{0011} & \textbf{0012} & \textbf{0019} & \textbf{0020} & \textbf{AP (per class)} \\ \hline
\multicolumn{1}{c|}{0} & \multicolumn{1}{l|}{instruction booklet} & 62.00\% & 38.78\% & 42.97\% & 63.75\% & 29.84\% & 38.25\% & 47.65\% & 46.18\% \\
\multicolumn{1}{c|}{1} & \multicolumn{1}{l|}{gray\_angled\_perforated\_bar} & 9.55\% & 18.81\% & 14.72\% & 2.17\% & 16.42\% & 0\% & 6.89\% & 9.79\% \\
\multicolumn{1}{c|}{2} & \multicolumn{1}{l|}{partial\_model} & 35.68\% & 31.74\% & 35.82\% & 42.55\% & 32.16\% & 33.02\% & 43.80\% & 36.40\% \\
\multicolumn{1}{c|}{3} & \multicolumn{1}{l|}{white\_angled\_perforated\_bar} & 43.70\% & 39.86\% & 9.90\% & 45.32\% & 24.94\% & 16.35\% & 33.31\% & 30.48\% \\
\multicolumn{1}{c|}{4} & \multicolumn{1}{l|}{wrench} & // & // & // & 11.11\% & // & 10.43\% & // & 10.77\% \\
\multicolumn{1}{c|}{5} & \multicolumn{1}{l|}{screwdriver} & 61.82\% & 57.68\% & 68.57\% & 54.21\% & 57.14\% & 62.68\% & 61.37\% & 60.50\% \\
\multicolumn{1}{c|}{6} & \multicolumn{1}{l|}{gray\_perforated\_bar} & 19.36\% & 40.26\% & 30.89\% & 53.06\% & 29.68\% & 26.82\% & 15.76\% & 30.83\% \\
\multicolumn{1}{c|}{7} & \multicolumn{1}{l|}{wheels\_axle} & 11.37\% & 18.34\% & 04.63\% & 1.79\% & 31.61\% & 03.91\% & 04.35\% & 10.86\% \\
\multicolumn{1}{c|}{8} & \multicolumn{1}{l|}{red\_angled\_perforated\_bar} & 18.65\% & 01.57\% & 4.81\% & 00.09\% & 12.27\% & 05.98\% & 09.64\% & 07.57\% \\
\multicolumn{1}{c|}{9} & \multicolumn{1}{l|}{red\_perforated\_bar} & 23.35\% & 26.69\% & 34.72\% & 24.58\% & 20.70\% & 11.21\% & 17.91\% & 22.74\% \\
\multicolumn{1}{c|}{10} & \multicolumn{1}{l|}{rod} & 14.90\% & 07.40\% & 22.41\% & 19.73\% & 15.57\% & 17.84\% & 14.04\% & 15.98\% \\
\multicolumn{1}{c|}{11} & \multicolumn{1}{l|}{handlebar} & 44.39\% & 36.31\% & 28.79\% & 26.92\% & 12.50\% & 27.27\% & 52.48\% & 32.67\% \\
\multicolumn{1}{c|}{12} & \multicolumn{1}{l|}{screw} & 48.64\% & 42.87\% & 40.00\% & 16.96\% & 44.99\% & 43.88\% & 35.35\% & 38.96\% \\
\multicolumn{1}{c|}{13} & \multicolumn{1}{l|}{tire} & 45.93\% & 71.68\% & 63.09\% & 89.01\% & 37.83\% & 39.69\% & 65.15\% & 58.91\% \\
\multicolumn{1}{c|}{14} & \multicolumn{1}{l|}{rim} & 45.10\% & 35.71\% & 42.57\% & 59.26\% & 22.28\% & 90.00\% & 57.54\% & 50.35\% \\
\multicolumn{1}{c|}{15} & \multicolumn{1}{l|}{washer} & 31.52\% & 39.39\% & 19.00\% & 19.57\% & 53.43\% & 44.45\% & 09.06\% & 30.92\% \\
\multicolumn{1}{c|}{16} & \multicolumn{1}{l|}{red\_perforated\_junction\_bar} & 19.28\% & 13.51\% & 07.55\% & 30.74\% & 28.63\% & 22.02\% & 16.89\% & 19.80\% \\
\multicolumn{1}{c|}{17} & \multicolumn{1}{l|}{red\_4\_perforated\_junction\_bar} & 24.20\% & 43.50\% & 39.11\% & 85.71\% & 44.23\% & 28.37\% & 20.62\% & 40.82\% \\
\multicolumn{1}{c|}{18} & \multicolumn{1}{l|}{bolt} & 33.14\% & 33.61\% & 11.29\% & 17.16\% & 28.46\% & 21.31\% & 19.12\% & 23.44\% \\
\multicolumn{1}{c|}{19} & \multicolumn{1}{l|}{roller} & 09.93\% & 40.50\% & 28.15\% & 5.76\% & 0.23\% & 18.20\% & 09.36\% & 16.02\% \\ \hline
\multicolumn{1}{l}{} &  & \multicolumn{1}{l}{} & \multicolumn{1}{l}{} & \multicolumn{1}{l}{} & \multicolumn{1}{l}{} & \multicolumn{1}{l}{} & \multicolumn{1}{l}{} & \multicolumn{1}{l}{} & \multicolumn{1}{l}{} \\ \cline{2-10} 
\multicolumn{1}{l}{} & \multicolumn{1}{c|}{\textbf{mAP (per video)}} & 31.71\% & 33.59\% & 28.89\% & 33.47\% & 28.57\% & 28.08\% & 28.44\% & \textbf{30.39\%}
\end{tabular}%
}
\caption{Baseline results for the \textit{active} object recognition task. We report the AP values for each class which are the averages of the AP values for each class of the Test videos. In the last column, we report the mAP per class, which is the average mAP of the Test videos.}
\label{tab:active_rec}
\end{table*}

\subsection {EHOI Detection}
For the ``SlowFast + Faster-RCNN'' baseline, we trained SlowFast network to recognize the $12$ verb classes of the MECCANO dataset using the same settings as the ones considered for the action recognition task. We trained the network for $40$ epochs and obtained a verb recognition Top-1 accuracy of $58.04\%$ on the Test set. 
For the object detector component, we used the same model trained for the active object recognition task. 

For the ``human-branch'' of the  ``InteractNet'' model, we used the Hand-Object Detector~\cite{Hands_in_contact_Shan20} to detect hands in the scene. The object detector trained for active object recognition has been used for the ``object-branch''. 
The MLPs used to predict the verb class form the appearance of hands and active objects are composed by an input linear layer (e.g., 1024-d for the hands MLP and 784-d for the objects one), a ReLU activation function and an output linear layer (e.g., 12-d for both MLPs). We fused by late fusion the output probability distributions of verbs obtained from the two MLPs (hands and objects) to predict the final verb of the EHOI. We jointly trained the MLPs for $50K$ iterations on an Nvidia V100 GPU, using a batch size of $28$ and a learning rate of $0.0001$.

In ``InteractNet + Context'', we added a third MLP which predicts the verb class based on context features. The context MLP has the same architecture of the others MLPs (hands and objects) except the input linear layer which is 640-d. In this case, we jointly trained the three MLPs (hands, objects and context) for $50K$ iterations on a TitanX GPU with a batch size equal to $18$ and the learning rate equal to $0.0001$.
The outputs of the three MLPs are hence fused by late fusion. \\

\section{Additional Results}
\label{ref:qualitative}
\begin{figure*}[t]
\centering
\includegraphics[width=\textwidth]{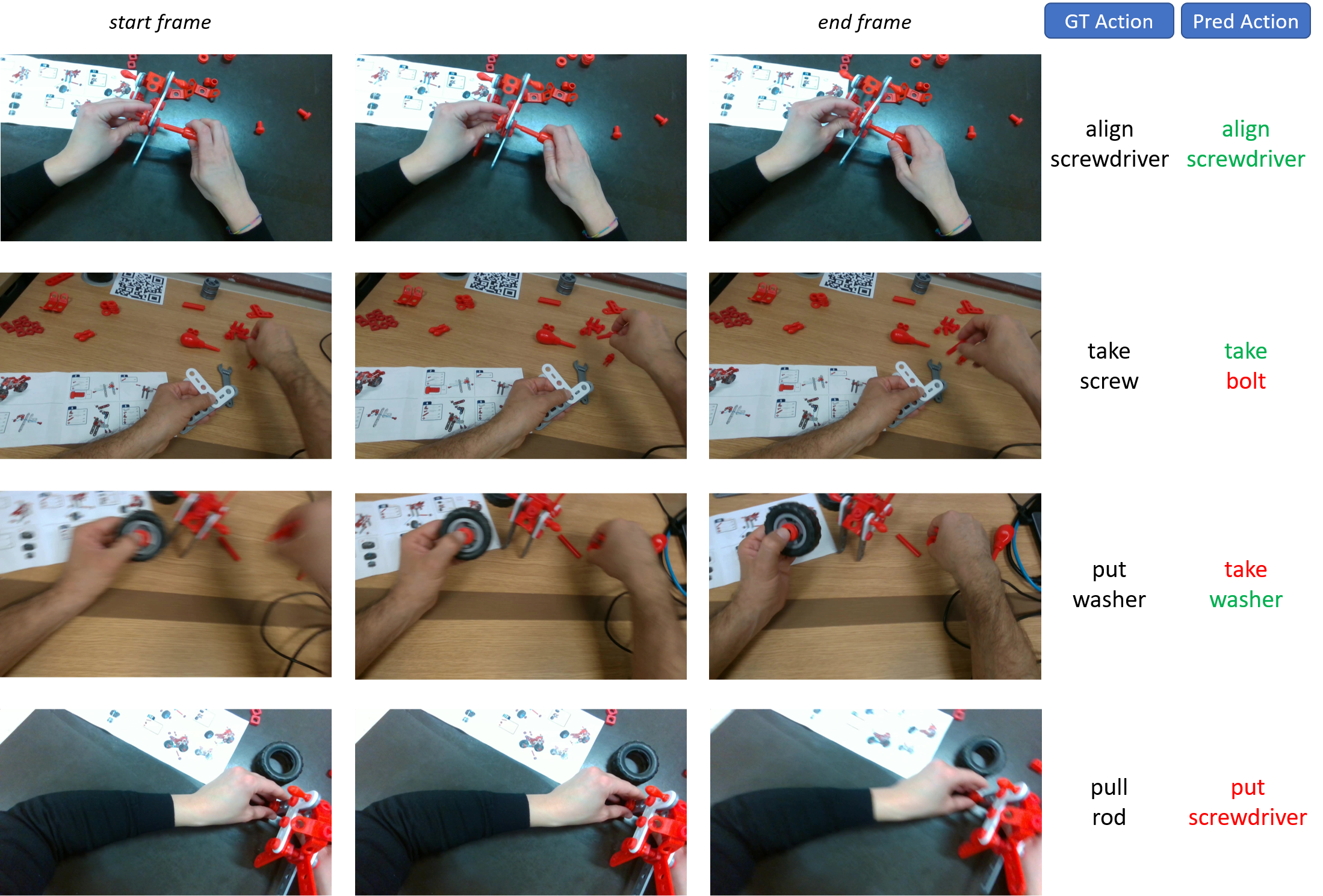}
\caption{Qualitative results for the action recognition task. Correct predictions are in green while wrong predictions are in red.}
\label{fig:actions_rec}
\end{figure*}

Figure~\ref{fig:actions_rec} shows some qualitative results of the SlowFast baseline. Note that, in the second and third example, the method predicts correctly only the verb or the object.

\begin{figure*}[t]
	\centering
	\includegraphics[width=\textwidth]{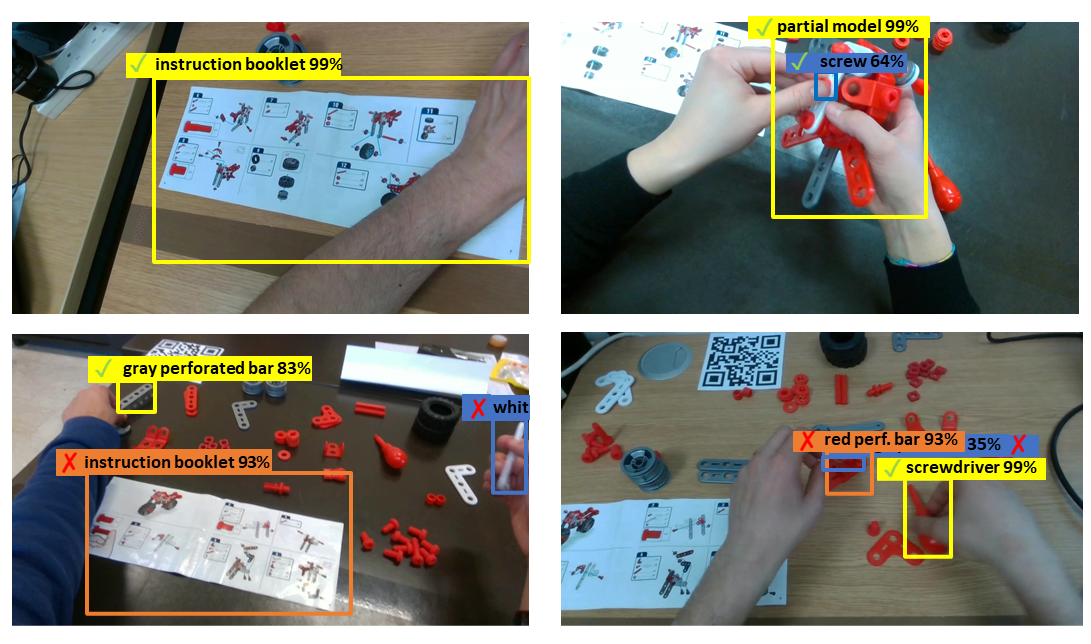}
	\caption{Qualitative results for the \textit{active} object recognition task.}
	\label{fig:active_objects_rec}
\end{figure*}

Table~\ref{tab:active_rec} reports the results obtained with the baseline in the \textit{Active} Object Recognition task.
We report the AP values for each class considering all the videos belonging to the test set of the MECCANO dataset. The last column shows the average of the AP values for each class and the last row reports the mAP values for each test video.
Figure~\ref{fig:active_objects_rec} reports some qualitative results for this task. In particular, in the first row, we report the correct \textit{active} object predictions, while in the second row we report two examples of wrong predictions. In the wrong predictions, the right \textit{active} object is recognized but other \textit{passive} objects are wrongly detected and recognized as \textit{active} (e.g., instruction booklet in the example bottom-left or the red bars in the example bottom-right of Figure~\ref{fig:active_objects_rec}).

\balance
{\small
\bibliographystyle{ieee_fullname}
\bibliography{egbibAcronims}

\begin{thebibliography}{10}\itemsep=-1pt

\bibitem{Epson_moverio}
Epson moverio bt 300.
\newblock
  \url{https://www.epson.eu/products/see-through-mobile-viewer/moverio-bt-300}.

\bibitem{Holo2}
Microsoft hololens 2.
\newblock \url{https://www.microsoft.com/en-us/hololens}.

\bibitem{Vuzix}
Vuzix blade.
\newblock \url{https://www.vuzix.com/products/blade-smart-glasses}.

\bibitem{Lending_Hand_Bambach_15}
S. {Bambach}, S. {Lee}, D.~J. {Crandall}, and C. {Yu}.
\newblock Lending a hand: Detecting hands and recognizing activities in complex
  egocentric interactions.
\newblock In {\em ICCV}, pages 1949--1957, 2015.

\bibitem{Cai2016UnderstandingHM}
Minjie Cai, Kris~M. Kitani, and Yoichi Sato.
\newblock Understanding hand-object manipulation with grasp types and object
  attributes.
\newblock In {\em Robotics: Science and Systems}, 2016.

\bibitem{Kinetics_Carreira2019ASN}
J. Carreira, Eric Noland, Chloe Hillier, and Andrew Zisserman.
\newblock A short note on the kinetics-700 human action dataset.
\newblock {\em ArXiv}, abs/1907.06987, 2019.

\bibitem{Carreira2017QuoVA}
Jo{\~a}o Carreira and Andrew Zisserman.
\newblock Quo vadis, action recognition? a new model and the kinetics dataset.
\newblock {\em CVPR}, pages 4724--4733, 2017.

\bibitem{HICO_Chao}
Y. {Chao}, Z. {Wang}, Y. {He}, J. {Wang}, and J. {Deng}.
\newblock Hico: A benchmark for recognizing human-object interactions in
  images.
\newblock In {\em ICCV}, pages 1017--1025, 2015.

\bibitem{Chao2018LearningTD}
Yu-Wei Chao, Yunfan Liu, Xieyang Liu, Huayi Zeng, and Jia Deng.
\newblock Learning to detect human-object interactions.
\newblock {\em WACV}, pages 381--389, 2018.

\bibitem{DeepVisionShield_Colombo19}
Sara Colombo, Yihyun Lim, and Federico Casalegno.
\newblock Deep vision shield: Assessing the use of hmd and wearable sensors in
  a smart safety device.
\newblock In {\em ACM PETRA}, 2019.

\bibitem{cucchiara2014visions}
Rita Cucchiara and Alberto Del~Bimbo.
\newblock Visions for augmented cultural heritage experience.
\newblock {\em IEEE MultiMedia}, 21(1):74--82, 2014.

\bibitem{Human_detection_Flow_Schmid_06}
Navneet Dalal, Bill Triggs, and Cordelia Schmid.
\newblock Human detection using oriented histograms of flow and appearance.
\newblock In {\em ECCV}, page 428–441, 2006.

\bibitem{Damen2018EPICKITCHENS}
D. Damen, H. Doughty, G.~M. Farinella, S. Fidler, A. Furnari, E. Kazakos, D.
  Moltisanti, J. Munro, T. Perrett, W. Price, and M. Wray.
\newblock Scaling egocentric vision: The epic-kitchens dataset.
\newblock In {\em ECCV}, 2018.

\bibitem{Damen2020Collection}
D. Damen, H. Doughty, G.~M. Farinella, S. Fidler, A. Furnari, E. Kazakos, D.
  Moltisanti, J. Munro, T. Perrett, W. Price, and M. Wray.
\newblock The epic-kitchens dataset: Collection, challenges and baselines.
\newblock {\em IEEE TPAMI}, 2020.

\bibitem{Damen2020RESCALING}
D. Damen, H. Doughty, G.~M. Farinella, A. Furnari, J. Ma, E. Kazakos, D.
  Moltisanti, J. Munro, T. Perrett, W. Price, and M. Wray.
\newblock Rescaling egocentric vision.
\newblock {\em CoRR}, abs/2006.13256, 2020.

\bibitem{You-Do_Damen_14}
Dima Damen, Teesid Leelasawassuk, Osian Haines, Andrew Calway, and Walterio
  Mayol-Cuevas.
\newblock You-do, i-learn: Discovering task relevant objects and their modes of
  interaction from multi-user egocentric video.
\newblock In {\em BMVC}, 2014.

\bibitem{Torre2009CMU-MMAC}
Fernando de~la Torre, Jessica~K. Hodgins, Javier Montano, and Sergio Valcarcel.
\newblock Detailed human data acquisition of kitchen activities: the
  cmu-multimodal activity database (cmu-mmac).
\newblock In {\em CHI 2009 Workshop. Developing Shared Home Behavior Datasets
  to Advance HCI and Ubiquitous Computing Research}, 2009.

\bibitem{dutta2019vgg}
Abhishek Dutta and Andrew Zisserman.
\newblock The {VIA} annotation software for images, audio and video.
\newblock In {\em Proceedings of the 27th ACM International Conference on
  Multimedia}, MM '19, New York, NY, USA, 2019. ACM.

\bibitem{PascalVOC_Zisserman_15}
Mark Everingham, S.~M. Eslami, Luc Gool, Christopher~K. Williams, John Winn,
  and Andrew Zisserman.
\newblock The pascal visual object classes challenge: A retrospective.
\newblock {\em Int. J. Comput. Vision}, 111(1):98–136, Jan. 2015.

\bibitem{caba2015activitynet}
Bernard~Ghanem Fabian Caba~Heilbron, Victor~Escorcia and Juan~Carlos Niebles.
\newblock Activitynet: A large-scale video benchmark for human activity
  understanding.
\newblock In {\em CVPR}, pages 961--970, 2015.

\bibitem{fan2020pyslowfast}
Haoqi Fan, Yanghao Li, Bo Xiong, Wan-Yen Lo, and Christoph Feichtenhofer.
\newblock Pyslowfast.
\newblock \url{https://github.com/facebookresearch/slowfast}, 2020.

\bibitem{affordance_Fang18}
K. {Fang}, T. {Wu}, D. {Yang}, S. {Savarese}, and J.~J. {Lim}.
\newblock Demo2vec: Reasoning object affordances from online videos.
\newblock In {\em CVPR}, pages 2139--2147, 2018.

\bibitem{vedi2019}
G.~M. Farinella, G. Signorello, S. Battiato, A. Furnari, F. Ragusa, R.
  Leonardi, E. Ragusa, E. Scuderi, A. Lopes, L. Santo, and M. Samarotto.
\newblock Vedi: Vision exploitation for data interpretation.
\newblock In {\em ICIAP}, 2019.

\bibitem{feichtenhofer2018slowfast}
Christoph Feichtenhofer, Haoqi Fan, Jitendra Malik, and Kaiming He.
\newblock Slowfast networks for video recognition.
\newblock In {\em ICCV}, pages 6202--6211, 2018.

\bibitem{Spatiotemporal_residual_action}
Christoph Feichtenhofer, Axel Pinz, and Richard~P. Wildes.
\newblock Spatiotemporal residual networks for video action recognition.
\newblock In {\em NeurIPS}, NIPS’16, page 3476–3484, Red Hook, NY, USA,
  2016. Curran Associates Inc.

\bibitem{Two-Stream_Zisserman}
Christoph Feichtenhofer, Axel Pinz, and Andrew Zisserman.
\newblock Convolutional two-stream network fusion for video action recognition.
\newblock In {\em CVPR}, 2016.

\bibitem{girshick2015fast}
Ross Girshick.
\newblock Fast {R-CNN}.
\newblock In {\em ICCV}, 2015.

\bibitem{girshick2014rich}
Ross Girshick, Jeff Donahue, Trevor Darrell, and Jitendra Malik.
\newblock Rich feature hierarchies for accurate object detection and semantic
  segmentation.
\newblock In {\em CVPR}, 2014.

\bibitem{Gkioxari2018DetectingAR}
Georgia Gkioxari, Ross~B. Girshick, Piotr Doll{\'a}r, and Kaiming He.
\newblock Detecting and recognizing human-object interactions.
\newblock {\em CVPR}, pages 8359--8367, 2018.

\bibitem{Something_Something_Goyal}
R. {Goyal}, S.~E. {Kahou}, V. {Michalski}, J. {Materzynska}, S. {Westphal}, H.
  {Kim}, V. {Haenel}, I. {Fruend}, P. {Yianilos}, M. {Mueller-Freitag}, F.
  {Hoppe}, C. {Thurau}, I. {Bax}, and R. {Memisevic}.
\newblock The “something something” video database for learning and
  evaluating visual common sense.
\newblock In {\em ICCV}, pages 5843--5851, 2017.

\bibitem{HOI_Gupta_09}
A. {Gupta}, A. {Kembhavi}, and L.~S. {Davis}.
\newblock Observing human-object interactions: Using spatial and functional
  compatibility for recognition.
\newblock {\em IEEE TPAMI}, 31(10), 2009.

\bibitem{Gupta2015VisualSR}
Saurabh Gupta and Jitendra Malik.
\newblock Visual semantic role labeling.
\newblock {\em ArXiv}, abs/1505.04474, 2015.

\bibitem{He2015_ResNet}
Kaiming He, Xiangyu Zhang, Shaoqing Ren, and Jian Sun.
\newblock Deep residual learning for image recognition.
\newblock {\em arXiv preprint arXiv:1512.03385}, 2015.

\bibitem{Kinetics_2017}
W. Kay, J. Carreira, K. Simonyan, Brian Zhang, Chloe Hillier, Sudheendra
  Vijayanarasimhan, F. Viola, T. Green, T. Back, A. Natsev, Mustafa Suleyman,
  and Andrew Zisserman.
\newblock The kinetics human action video dataset.
\newblock {\em ArXiv}, abs/1705.06950, 2017.

\bibitem{kazakos2019TBN}
Evangelos Kazakos, Arsha Nagrani, Andrew Zisserman, and Dima Damen.
\newblock Epic-fusion: Audio-visual temporal binding for egocentric action
  recognition.
\newblock In {\em ICCV}, 2019.

\bibitem{Imagenet}
Alex Krizhevsky, Ilya Sutskever, and Geoffrey~E Hinton.
\newblock Imagenet classification with deep convolutional neural networks.
\newblock In F. Pereira, C.~J.~C. Burges, L. Bottou, and K.~Q. Weinberger,
  editors, {\em NeurIPS}, pages 1097--1105. 2012.

\bibitem{Learning_actions_movies_Laptev}
Ivan Laptev, Marcin Marszalek, Cordelia Schmid, and Benjamin Rozenfeld.
\newblock Learning realistic human actions from movies.
\newblock In {\em CVPR}. IEEE Computer Society, 2008.

\bibitem{Li2018_EGTEA-GAZE+}
Yin Li, Miao Liu, and James~M. Rehg.
\newblock In the eye of beholder: Joint learning of gaze and actions in first
  person video.
\newblock In {\em ECCV}, 2018.

\bibitem{PPDM_liao2019}
Yue Liao, Si Liu, Fei Wang, Yanjie Chen, Chen Qian, and Jiashi Feng.
\newblock {PPDM}: Parallel point detection and matching for real-time
  human-object interaction detection.
\newblock In {\em CVPR}, 2020.

\bibitem{TSM_2019}
J. {Lin}, C. {Gan}, and S. {Han}.
\newblock Tsm: Temporal shift module for efficient video understanding.
\newblock In {\em ICCV}, pages 7082--7092, 2019.

\bibitem{lin2014COCO}
Tsung-Yi Lin, Michael Maire, Serge Belongie, Lubomir Bourdev, Ross Girshick,
  James Hays, Pietro Perona, Deva Ramanan, C.~Lawrence Zitnick, and Piotr
  Dollár.
\newblock Microsoft coco: Common objects in context, 2014.
\newblock cite arxiv:1405.0312.

\bibitem{Ma_2016_CVPR}
Minghuang Ma, Haoqi Fan, and Kris~M. Kitani.
\newblock Going deeper into first-person activity recognition.
\newblock In {\em CVPR}, 2016.

\bibitem{Hotspots_Grauman19}
T. {Nagarajan}, C. {Feichtenhofer}, and K. {Grauman}.
\newblock Grounded human-object interaction hotspots from video.
\newblock In {\em ICCV}, pages 8687--8696, 2019.

\bibitem{Nagarajan2020EGOTOPOEA}
Tushar Nagarajan, Yanghao Li, Christoph Feichtenhofer, and Kristen Grauman.
\newblock Ego-topo: Environment affordances from egocentric video.
\newblock {\em ArXiv}, abs/2001.04583, 2020.

\bibitem{ELAN}
The Language~Archive Nijmegen: Max Planck Institute~for Psycholinguistics.
\newblock Elan (version 5.9) [computer software].
\newblock 2020.

\bibitem{Ortis2017OrganizingEV}
A. Ortis, G. Farinella, V. D'Amico, Luca Addesso, Giovanni Torrisi, and S.
  Battiato.
\newblock Organizing egocentric videos of daily living activities.
\newblock {\em Pattern Recognition}, 72, 2017.

\bibitem{Ramanan_12_ADL}
H. {Pirsiavash} and D. {Ramanan}.
\newblock Detecting activities of daily living in first-person camera views.
\newblock In {\em CVPR}, 2012.

\bibitem{Qi2018LearningHI}
Siyuan Qi, Wenguan Wang, Baoxiong Jia, Jianbing Shen, and Song-Chun Zhu.
\newblock Learning human-object interactions by graph parsing neural networks.
\newblock {\em ArXiv}, abs/1808.07962, 2018.

\bibitem{Learning_spatiotemporal_pseudo}
Zhaofan Qiu, Ting Yao, and Tao Mei.
\newblock Learning spatio-temporal representation with pseudo-3d residual
  networks.
\newblock pages 5534--5542, 10 2017.

\bibitem{RagusaPRL}
F. Ragusa, A. Furnari, S. Battiato, G. Signorello, and G.~M. Farinella.
\newblock {EGO-CH}: Dataset and fundamental tasks for visitors behavioral
  understanding using egocentric vision.
\newblock {\em Pattern Recognition Letters}, 2020.

\bibitem{yolov3}
Joseph Redmon and Ali Farhadi.
\newblock Yolov3: An incremental improvement.
\newblock {\em CoRR}, abs/1804.02767, 2018.

\bibitem{ren2015faster}
Shaoqing Ren, Kaiming He, Ross Girshick, and Jian Sun.
\newblock Faster {R-CNN}: Towards real-time object detection with region
  proposal networks.

\bibitem{Hands_in_contact_Shan20}
Dandan Shan, Jiaqi Geng, Michelle Shu, and David Fouhey.
\newblock Understanding human hands in contact at internet scale.
\newblock 2020.

\bibitem{Sigurdsson2018Charades}
Gunnar~A. Sigurdsson, Abhinav Gupta, C. Schmid, Ali Farhadi, and Alahari
  Karteek.
\newblock Actor and observer: Joint modeling of first and third-person videos.
\newblock {\em CVPR}, pages 7396--7404, 2018.

\bibitem{TwoStream_convolutional_action_Zisserman_14}
Karen Simonyan and Andrew Zisserman.
\newblock Two-stream convolutional networks for action recognition in videos.
\newblock In {\em NeurIPS}, 2014.

\bibitem{miss_actions_shapiro}
Bilge Soran, Ali Farhadi, and Linda Shapiro.
\newblock Generating notifications for missing actions: Don't forget to turn
  the lights off!
\newblock pages 4669--4677, 12 2015.

\bibitem{Sudhakaran_2019_CVPR}
Swathikiran Sudhakaran, Sergio Escalera, and Oswald Lanz.
\newblock Lsta: Long short-term attention for egocentric action recognition.
\newblock In {\em CVPR}, June 2019.

\bibitem{thu-read_17}
Y. {Tang}, Y. {Tian}, J. {Lu}, J. {Feng}, and J. {Zhou}.
\newblock Action recognition in rgb-d egocentric videos.
\newblock In {\em ICIP}, pages 3410--3414, 2017.

\bibitem{Conv_spatio-temporal_Taylor}
Graham~W. Taylor, Rob Fergus, Yann LeCun, and Christoph Bregler.
\newblock Convolutional learning of spatio-temporal features.
\newblock In {\em ECCV}, page 140–153, 2010.

\bibitem{Learning_spatio-temporal_Paluri}
D. {Tran}, L. {Bourdev}, R. {Fergus}, L. {Torresani}, and M. {Paluri}.
\newblock Learning spatiotemporal features with 3d convolutional networks.
\newblock In {\em ICCV}, pages 4489--4497, 2015.

\bibitem{closer_spatiotemp_action}
D. {Tran}, H. {Wang}, L. {Torresani}, J. {Ray}, Y. {LeCun}, and M. {Paluri}.
\newblock A closer look at spatiotemporal convolutions for action recognition.
\newblock In {\em CVPR}, pages 6450--6459, 2018.

\bibitem{Long-term_action_Schmid}
G. {Varol}, I. {Laptev}, and C. {Schmid}.
\newblock Long-term temporal convolutions for action recognition.
\newblock {\em IEEE TPAMI}, 40(6):1510--1517, 2018.

\bibitem{temporal_segNet}
Limin Wang, Yuanjun Xiong, Zhe Wang, Yu Qiao, Dahua Lin, Xiaoou Tang, and Luc
  Van~Gool.
\newblock Temporal segment networks: Towards good practices for deep action
  recognition.
\newblock volume 9912, 10 2016.

\bibitem{Wang_InteractionPoints_2020_CVPR}
Tiancai Wang, Tong Yang, Martin Danelljan, Fahad~Shahbaz Khan, Xiangyu Zhang,
  and Jian Sun.
\newblock Learning human-object interaction detection using interaction points.
\newblock In {\em CVPR}, June 2020.

\bibitem{wu2019detectron2}
Yuxin Wu, Alexander Kirillov, Francisco Massa, Wan-Yen Lo, and Ross Girshick.
\newblock Detectron2.
\newblock \url{https://github.com/facebookresearch/detectron2}, 2019.

\bibitem{rethinking_spatiotemporal}
Saining Xie, Chen Sun, Jonathan Huang, Zhuowen Tu, and Kevin Murphy.
\newblock Rethinking spatiotemporal feature learning for video understanding.
\newblock {\em CoRR}, abs/1712.04851, 2017.

\bibitem{HOI_Fei_Fei}
B. {Yao} and L. {Fei-Fei}.
\newblock Recognizing human-object interactions in still images by modeling the
  mutual context of objects and human poses.
\newblock {\em IEEE TPAMI}, 34(9):1691--1703, 2012.

\bibitem{Zhou2018TemporalRR}
Bolei Zhou, Alex Andonian, and Antonio Torralba.
\newblock Temporal relational reasoning in videos.
\newblock {\em ArXiv}, abs/1711.08496, 2018.

\bibitem{RPN_Zhou}
P. {Zhou} and M. {Chi}.
\newblock Relation parsing neural network for human-object interaction
  detection.
\newblock In {\em ICCV}, pages 843--851, 2019.

\end{thebibliography}
}

\end{document}